\documentclass[10pt,twocolumn,letterpaper]{article}

\usepackage{cvpr}              % To produce the CAMERA-READY version
\usepackage{graphicx}
\usepackage{amsmath}
\usepackage{amssymb}
\usepackage{times}
\usepackage{epsfig}
\usepackage{amsmath,amsfonts,bm}

\def\eqref#1{equation~\ref{#1}}
\def\1{\bm{1}}

\def\va{{\bm{a}}}

\def\vp{{\bm{p}}}

\def\vv{{\bm{v}}}

\def\vx{{\bm{x}}}

\def\mV{{\bm{V}}}

\def\mX{{\bm{X}}}
\def\mY{{\bm{Y}}}

\DeclareMathAlphabet{\mathsfit}{\encodingdefault}{\sfdefault}{m}{sl}
\SetMathAlphabet{\mathsfit}{bold}{\encodingdefault}{\sfdefault}{bx}{n}

\DeclareMathOperator*{\argmin}{arg\,min}

\usepackage{multirow, float, booktabs, boldline, hhline}
\usepackage{algorithm}
\usepackage{algpseudocode}
\usepackage{caption}
\usepackage{subcaption}
\usepackage{ctable}
\usepackage{diagbox}

\usepackage[pagebackref,breaklinks,colorlinks]{hyperref}

\usepackage[capitalize]{cleveref}
\crefname{section}{Sec.}{Secs.}
\Crefname{section}{Section}{Sections}
\Crefname{table}{Table}{Tables}
\crefname{table}{Tab.}{Tabs.}

\begin{document}

\title{Federated Zero-Shot Learning for Visual Recognition}

\author{Zhi Chen$^1$ \quad  Yadan Luo$^1$ \quad  Sen Wang$^1$ \quad Jingjing Li$^2$ \quad Zi Huang$^1$ \quad   \\ 
$^1$ The University of Queensland, Australia  \\ $^2$University of Electronic Science and Technology of China   \\ 
\{zhi.chen,y.luo,sen.wang\}@uq.edu.au \\ lijin117@yeah.net, huang@itee.uq.edu.au}
\maketitle

%%%%%%%%% ABSTRACTs
\begin{abstract}
Zero-shot learning is a learning regime that recognizes unseen classes by generalizing the visual-semantic relationship learned from the seen classes. To obtain an effective ZSL model, one may resort to curating training samples from multiple sources, which may inevitably raise the privacy concerns about data sharing across different organizations. In this paper, we propose a novel Federated Zero-Shot Learning (\texttt{FedZSL}) framework, which learns a central model from the decentralized data residing on edge devices. To better generalize to previously unseen classes, FedZSL allows the training data on each device sampled from the non-overlapping classes, which are far from the \textit{i.i.d.} that traditional federated learning commonly assumes. We identify two key challenges in our FedZSL protocol: 1) the trained models are prone to be biased to the locally observed classes, thus failing to generalize to the unseen classes and/or seen classes appeared on other devices; 2) as each category in the training data comes from a single source, the central model is highly vulnerable to model replacement (backdoor) attacks. To address these issues, we propose three local objectives for visual-semantic alignment and cross-device alignment through relation distillation, which leverages the normalized class-wise covariance to regularize the consistency of the prediction logits across devices. To defend against the backdoor attacks, a feature magnitude defending technique is proposed. As malicious samples are less correlated to the given semantic attributes, the visual features of low magnitude will be discarded to stabilize model updates. The effectiveness and robustness of FedZSL are demonstrated by extensive experiments conducted on three zero-shot benchmark datasets.

\end{abstract}

%%%%%%%%% BODY TEXT
\section{Introduction}

\begin{figure}
    \centering
    \includegraphics[width=1\linewidth]{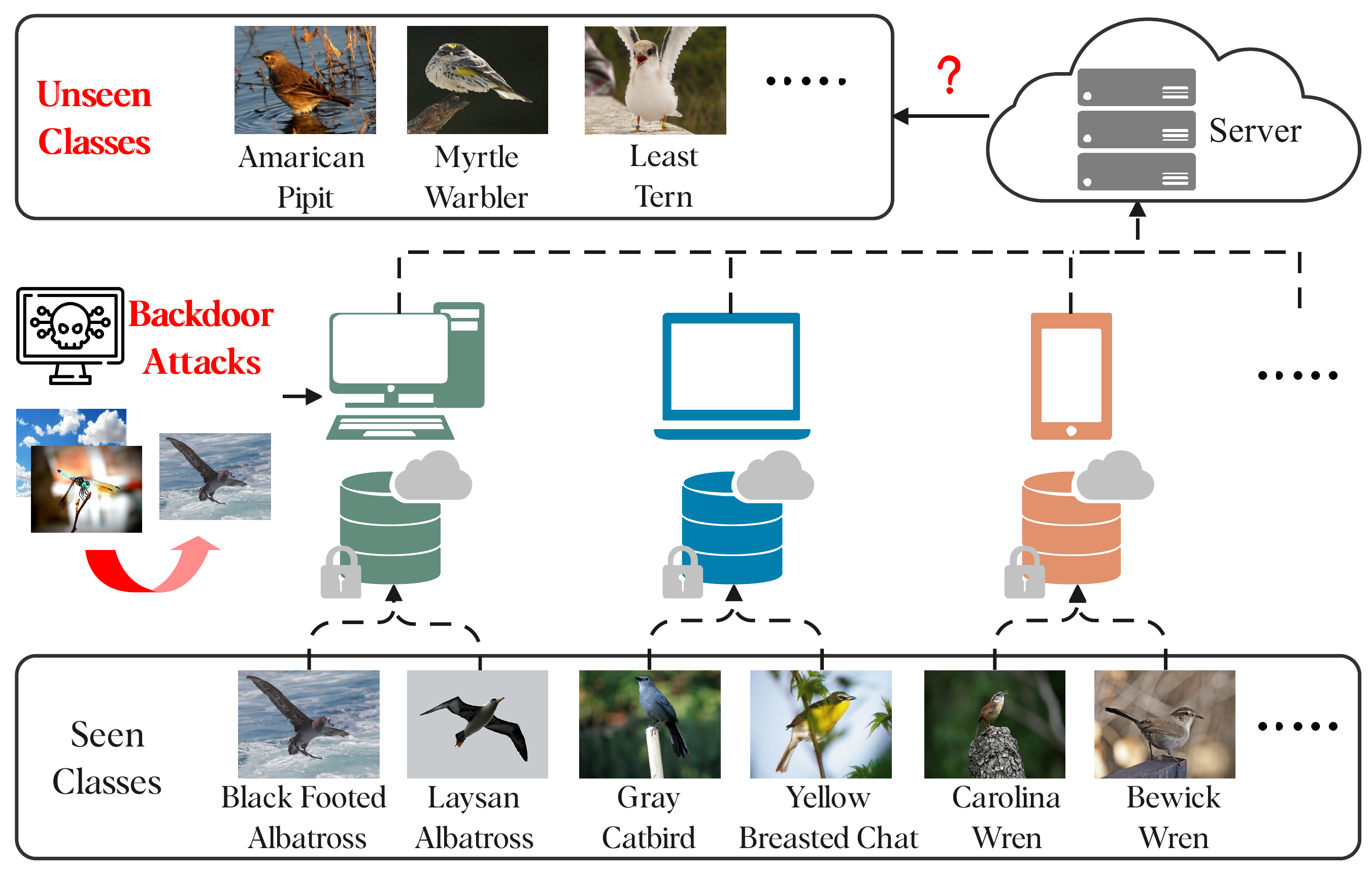}
    % \vspace{-15pt}
    \caption{An illustration of a novel Federated Zero-Shot Learning (\texttt{FedZSL}) framework, which aims to learn a central model from multiple participants. The local data on edge devices follows the  partial class-conditional distribution (\textit{p.c.c.d}), which is far from \textit{i.i.d.} that conventional federated learning models assume.}
    % that possess unique sets of seen classes such that the federated model can directly generalize to both seen classes and unseen classes.}
    \label{fig:intro}
    % \vspace{-15pt}
\end{figure}

With the ever-growing number of novel concepts, the ability to faithfully recognize instances from the previously \textit{seen} and \textit{unseen} classes is highly desired in modern vision systems. In response to this demand, generalized zero-shot learning (GZSL) \cite{zhang2017learning,xian2018feature,liu2018generalized,schonfeld2019generalized} has been emerging as a feasible solution. GZSL approaches learn to classify \textit{seen} classes (\textit{i.e.,} appeared during training and test) and \textit{unseen} classes (\textit{i.e.,} only encountered in testing phase) based on their semantic descriptions without requiring any exemplar of novel classes. The missing piece in the visual-semantic mapping for unseen classes can be inferred by constructing an intermediate semantic space \cite{akata2015label}, wherein the knowledge from seen classes is capable of being transferred to new concepts. To obtain an effective classification model, the mainstream GZSL workflow commonly curates large-scale training data from multiple parties into a centralized server. Nevertheless, such a centralized training paradigm may be infeasible in practice. Due to increasing concerns about data privacy and security, it is highly restricted to share the training data across different organizations. 

\begin{figure*}
    \centering
    \includegraphics[width=0.9\linewidth]{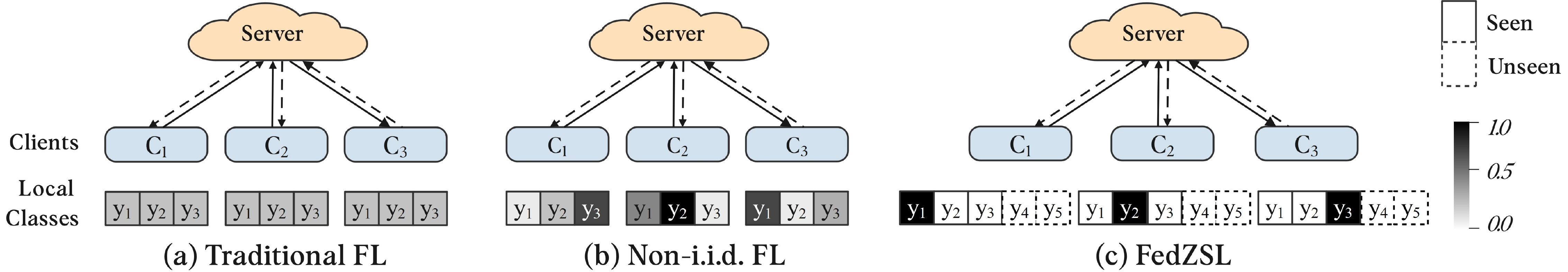}
    % \vspace{-8pt}
    \caption{Data distribution comparisons between traditional FL with \textit{i.i.d. setting}, \textit{non-i.i.d.} FL and our proposed \texttt{FedZSL} with \textit{p.c.c.d.}}
    \label{fig:fedzsl}
    % \vspace{-15pt}
\end{figure*}

In this paper, we resort to \textit{federated learning} (FL) \cite{mcmahan2017communication,li2021ditto,jiang2019improving}, a privacy-preserving framework for learning models from multiple decentralized data sources with no exchange of local training data. As shown in Fig. \ref{fig:fedzsl}(\textcolor{red}{a}) and Fig. \ref{fig:fedzsl}(\textcolor{red}{b}), traditional FL approaches assume all participants possess an identical set of classes with either \textit{i.i.d.}  or \textit{non-i.i.d.}  sampled data. Complementary to previous work, our study investigates a more practical setting namely \textit{federated zero-shot learning} (\texttt{FedZSL}), where multiple parties exclusively hold the training data from non-overlapping classes, as depicted in Fig. \ref{fig:fedzsl}(\textcolor{red}{c}). Here, we refer to this distribution that the local training data follows as a partial class-conditional distribution (\textit{p.c.c.d.}). We consider that the FedZSL is also more flexible and scalable, since it will enforce the global model to learn how to quickly adapt to the new seen classes when new participants join in the FL cluster. As shown in Fig. \ref{fig:intro}, by aggregating the local updates from various edge devices, FedZSL aims to find a global model at the server-side that can accurately recognize all the seen classes (\textit{e.g.,} black footed albatross) that are distributed locally. Meanwhile, the learned model is expected to generalize well to the unseen classes (\textit{e.g.,} American pipit) that are not presented in the training stage.

Despite the FedZSL protocol being ideal for zero-shot recognition tasks, it poses a significant challenge on system effectiveness and robustness, mainly from two perspectives: (1) the \textit{p.c.c.d.} data makes FedZSL the worst case of \textit{non-i.i.d.} FL. With no access to visual samples of any other classes, the trained models are prone to be biased to the classes that are locally available, thus failing to generalize to the unseen classes and/or seen classes appeared in other devices; (2) as each class in the training data comes from a single source, the central model is highly vulnerable to model replacement (backdoor) attacks \cite{bagdasaryan2020backdoor,ozdayi2021defending,chen2017targeted}. The erroneous data uploaded by careless or malicious participants involved in training process may lead to performance degradation of both local and global models, which becomes more severe as the communication round goes up.

To address these challenges, in this work, we investigate how to learn an effective and robust FedZSL model that is able to faithfully classify the seen and unseen classes while defending against multiple types of backdoor attacks. In particular, three local objectives are designed for visual-semantic alignment and cross-device alignment through relation distillation, which leverages the normalized class-wise covariance estimated by Graphical Lasso to regularize the consistency of the prediction across devices. In addition, to defend against the backdoor attacks, a \textit{feature magnitude defense} (FMD) method is proposed to detect and remove the potential malicious samples by examining the correlation of semantic attributes in the visual representation of each instance. As malicious samples typically have irrelevant backgrounds, styles and objects, the associated visual features are of a low magnitude. Hence, the visual features with low magnitude will be discarded, resulting in stable updates of local models.
The main contributions of this work are summarized as follows:

\begin{itemize}
    \item This work is the very first attempt to engage multiple parties to collaboratively learn a unified zero-shot model with partial class-conditional distribution (\textit{p.c.c.d.}). A novel and practical setting of FedZSL is formulated, which is complementary to all existing FL frameworks.
    \item To deal with the lack of global semantic information in local model training, we propose a \textit{cross-device alignment via relation distillation} (CARD) training strategy, which reconciles the visual-semantic relationship learning among local models that are trained with exclusively owned classes.
    \item We improve the robustness of FedZSL by preventing backdoor attacks with the proposed \textit{feature magnitude defense} (FMD). Three different types of backdoor attacks on backgrounds, styles, and objects are studied. 
    \item Extensive experiments conducted on three benchmark zero-shot learning datasets evidence the effectiveness and the robustness of the proposed FedZSL.% framework. 
    % Source code is made available in supplementary material.
\end{itemize}

% We propose to progressively integrate and adapt.

% "In the 21st century, we were confronting many big data breaches that necessitate governments, organizations, and companies to give a reconsideration of privacy. In contrast to that, almost of breakthroughs in Machine Learning come from learning techniques which requires a large amount of training data. Besides, research institutions often use and share data containing sensitive or confidential information about individuals. Improper disclosure of such data can have adverse consequences for a data subject’s private information, or even lead to civil liability or bodily harm."

% "In short, Differential Privacy permits:
% — Companies access a large number of sensitive data for researching and business without privacy breach.
% — Research institutions can develop differential privacy technology to automate privacy processes within cloud-sharing communities across countries. Thus, they could protect the privacy of users and resolve data sharing problem."

%-------------------------------------------------------------------------
\section{Related Work}
\subsection{Zero-Shot Learning}
Zero-shot Learning \cite{romera2015embarrassingly,xian2017zero,kodirov2017semantic} addresses a challenging problem where the test set contains additional classes not presented during training. A standard solution is to leverage intermediate class-level semantic representations (\textit{e.g.,} attribute annotations \cite{lampert2013attribute} or natural language descriptions \cite{elhoseiny2013write}) to bridge the seen and unseen classes \cite{akata2015label,li2017zero}. There are roughly two streams of methods to learn the visual-semantic relationships:  
% Zero-shot learning can be roughly categorized as two branches: 
embedding-based methods \cite{xu2020attribute,palatucci2009zero,norouzi2013zero,zhang2017learning} and generative methods \cite{xian2018feature,verma2018generalized,zhu2018generative,han2021contrastive}. The former group learns to project the visual and semantic information from all classes to the same space, and use the learned mapping to infer the class attributes for the unseen samples at the test time. The latter group needs to firstly train a generative model from semantic information to visual features, and then leverage the trained generative model to hallucinate the potential visual features for the unseen classes. As generative approaches generally involve large-scale data synthesis and heavy computation, it will lead to high overhead costs on edge devices if extended to a federated learning scheme. Hence, we follow the embedding-based workflow. 
% Through the observations in Section \ref{sec:exp}, we find that learning from more seen classes is beneficial to improve the visual-semantic generalization ability. However, in the real world, most seen classes are proprietary and not shared publicly due to privacy or confidentiality concerns. To utilize the private seen classes, in this paper, we study federated zero-shot learning (\texttt{FedZSL}) that learns from on-device data in partial class-conditional distribution as shown in Figure \ref{fig:fedzsl}.

%-------------------------------------------------------------------------

\subsection{Federated Learning (FL)}
Federated learning \cite{mcmahan2017communication,mohassel2017secureml} is a decentralized learning protocol that enables multiple participants to collaboratively learn a unified model without sharing the local data, which provides a promising privacy-preserving solution.  Researchers in this area have been dedicated to improving efficiency and effectiveness (\textit{e.g.,} strategies for dealing with \textit{non-i.i.d.} data \cite{stich2018local,yu2018parallel,li2019convergence}), preserving the privacy of user data \cite{bogdanov2012deploying,agrawal2019quotient}, ensuring fairness and addressing sources of bias \cite{li2019fair,jiang2019improving}, and addressing system challenges \cite{sheller2020federated,bonawitz2019towards}. 

Note that our proposed FedZSL is different from existing zero-shot related methods in federated learning \cite{hao2021towards,zhang2021fedzkt,gudur_interspeech21}, which are incapable of directly generalizing the federated model to unseen classes. Instead, they only consider data or model heterogeneity problems. Specifically, Fed-ZD \cite{hao2021towards} considers improving model fairness on under-representative  classes that only partial clients hold. They propose to perform data augmentation for those under-representative classes. FedZKT \cite{zhang2021fedzkt} is proposed to solve the model heterogeneity by distilling the knowledge from heterogeneous local models. 
Fed-NCAC \cite{gudur_interspeech21} is inspired by the data impression technique \cite{nayak2019zero} that adapts the current model to emerging new classes. However, the adaption is based on training samples of new classes.
This paper considers FedZSL with partial class-conditional distribution (\textit{p.c.c.d.}) data, which intrinsically solves the zero-shot learning problem in a federated learning scenario.

%-------------------------------------------------------------------------

\subsection{Backdoor Attacks in FL}
Adversarial attacks can be roughly categorized as untargeted and backdoor (targeted) attacks. In untargeted attacks \cite{blanchard2017machine,damaskinos2018asynchronous}, the malicious participants aim to corrupt the model, resulting in inferior performance on the primary tasks. In contrast, backdoor (targeted) attacks \cite{chen2017targeted} are targeted explicitly on certain sub-tasks and make them work with backdoored functionalities. Meanwhile, these backdoor functionalities should not affect the performance of the primary tasks. 
% For example, in the autonomous driving scenario, a malicious attacker may inject images of a specific advertising banner to the class of the stop sign, controlling cars to stop in front of the advertising banners.
Bagdasaryan \textit{et al.} \cite{bagdasaryan2020backdoor} showed federated learning is generally vulnerable to backdoors as it is impossible to ensure none of the participants are malicious. 
% Also, through experiments they demonstrate that simply tampering with training data does not work well against federated learning, as the server-side aggregation cancels out most of the backdoored model's contribution and can quickly forget the backdoor in a few rounds after submitting the malicious model. 
The model replacement attack is introduced to scale the backdoored model by a constant scaling factor to dominate the aggregation updates, thus preventing the backdoor functionalities from being canceled out. Sun \textit{et al.} \cite{sun2019can} leveraged differential privacy to defend against backdoor tasks by first clipping gradient updates and then adding Gaussian noise to the local model. Ditto \cite{li2021ditto} is proposed to improve both robustness and fairness in federated learning; interpolating between local and global models is proved to be effective. Ozdayi \textit{et al.} \cite{ozdayi2021defending} proposed to defend the backdoor functionalities by carefully adjusting the aggregation server's learning rate per dimension and round, based on the sign information of agents' updates. Unlike these sophisticated methods, in FedZSL, we propose a lightweight method with minimal changes to FL protocol to defend the backdoor attacks by inspecting the attribute presences in the input samples.

%-------------------------------------------------------------------------
\section{Federated Zero-Shot Learning}
This section will first formally introduce the \texttt{FedZSL} task, and then detail the global and local training procedure. Three different alignment objectives are applied to address the partial class-conditional distribution and misalignment of visual and semantic space. Finally, the backdoor defending with a feature magnitude mechanism is introduced.

%-------------------------------------------------------------------------
\subsection{Overview of FedZSL}
We consider a distributed system of $K$ clients $C_1, C_2,\dots, C_K$, each of which owns a local data source for training, \textit{i.e.,} $\mathcal{D}^{s} = \{\mathcal{D}^{s,1}, \mathcal{D}^{s,2}, \ldots, \mathcal{D}^{s,K}\}$. In particular, the $k$-th device has $N^k$ pairs of images with annotations, \textit{i.e.,} $\mathcal{D}^{s,k}=\{(\vx_{i}^{s,k}, y_{i}^{s,k})\}_{i=1}^{N^{k}}$, where only a part of seen classes are observable $y_{i}^{s,k}\in \mathcal{Y}^{s,k}$. Notably, the seen classes across the devices are non-overlapping, $\bigcap_{k \in [K]}\mathcal{Y}^{s,k}=\emptyset$ and $\bigcup_{k \in [K]} \mathcal{Y}^{s,k} = \mathcal{Y}^s$, where $|\mathcal{Y}^{s}|$ means the total number of the seen classes across $K$ devices. In FedZSL, we follow the standard setting \cite{mcmahan2017communication} that each client trains a local recognition model based on the local data, while a central server will collect the parameters periodically, aggregate them to update the global parameters for recognizing both seen classes $\mathcal{Y}^s$ and $\mathcal{Y}^u$. For briefty, we define $|\mathcal{Y}^{s}|+|\mathcal{Y}^{u}| = |\mathcal{Y}|$. To enable the parameter sharing between labels, the semantic information $\mathcal{A} = \{\va_{j} \}_{j=1}^{|\mathcal{Y}|}\in\mathbb{R}^{d_a\times |\mathcal{Y}|}$ is shared among all devices, where $d_a$ represents the attribute dimension.

Formally, we leverage the training data on $K$ devices $\mathcal{D}^{s} \triangleq \bigcup_{k \in [K]} \mathcal{D}^{s,k}$ to initiate a unified model $w$. The global learning objective is to solve:
\begin{equation}
\begin{aligned}
  \underset{w}{\min} \mathcal{L}(w) = \sum_{k=1}^{K} \dfrac{|\mathcal{Y}^{s,k}|}{|\mathcal{Y}^s|} \mathcal{L}^{k}(w),
\end{aligned}
\end{equation}
where $\mathcal{L}_{k}(w) = \mathop{\mathbb{E}}_{(\vx,y)\sim \mathcal{D}^{s,k}} [\ell^{k}(w;(\vx,y))]$ is the empirical loss of the client $C_{k}$. Different from the weighting scheme defined in FederatedAveraging (FedAvg) \cite{mcmahan2017communication}, a seen class ratio is utilized to measure the shared information available locally. We denote the model parameters at the round $t$ by $w_t$ and the $k$-th local model update by $\bigtriangleup w_{t}^k$. Therefore, the server will update the global model by aggregating $k$-th participant's local updates by:
\begin{equation}
\begin{aligned}
  w_{t+1} = w_t + \eta \frac{|\mathcal{Y}^{s,k}|\bigtriangleup w_t^k }{|\mathcal{Y}^{s}|},
 \label{modelupdate}
\end{aligned}
\end{equation}
where $\eta$ is the learning rate at the server side. The overall training procedure can be found in Algorithm \ref{alg}.

\subsection{Local Training}\label{sec:local}
During the local training procedure in the communication round $t$, the client $C_k$ receives the aggregated model weight $w_t$ and applies it to the local model $w_t^k$. Given an input image $\vx_i$, a pre-trained CNN backbone is leveraged as an image encoder $f(\cdot)$ to produce the visual features $\vv_i = f(\vx_i) \in \mathbb{R}^{d_v}$, where $d_v$ denotes the visual features' dimension. In contrast to conventional classification models having a fully connected layer atop the backbone to produce logits, ZSL methods usually add an attribute regression layer $g(\cdot): d_v \rightarrow d_a$ to calculate the presences of the semantic attributes from the visual features, $\widehat{\va}_i = g(f(\vx_i))\in\mathbb{R}^{d_a}$.

\noindent\textbf{Visual-Semantic Alignment.} For simplicity, we adopt a linear layer to map the visual features to semantic attributes. To determine the correct class that the predicted semantic attributes correspond to, the dot product operation is performed between the predicted semantic attributes and all possible semantic attributes to calculate the class logits. The semantic cross-entropy (SCE) loss is the objective to encourage the input images to have the highest compatibility score with their corresponding semantic attributes, which can be formulated as:
\begin{equation}
\begin{aligned}
  \ell_{sce} = - \sum_{\vx_i \in \mathcal{D}^{s,k}} \log \frac{\exp(\widehat{\va}_i^{T} \cdot \va_j)}{\sum_{\va_k \in \mathcal{A}} \exp (\widehat{\va}_i^{T} \cdot \va_{k})}.
\end{aligned}
\end{equation}
The attribute regression layer $g(\cdot)$ and the backbone $f(\cdot)$ are trained jointly.

\noindent\textbf{Cross-device Alignment.} To deal with \textit{p.c.c.d.} data distribution across devices, the idea of our solution is to align the class relationships in the local models across devices, so that we can learn a consistent visual space.
However, as the training samples that involve visual information are strictly preserved in local devices, it is elusive to align the visual space with the collaborative classes. We propose a \textit{cross-device alignment via relation distillation} (CARD) training strategy to align the visual space in different local models according to the class similarities in the semantic space.
Specifically, we start with constructing a class semantic similarity matrix. Graphical Lasso \cite{friedman2008sparse} is leveraged to estimate the sparse covariance $\Sigma\in\mathbb{R}^{|\mathcal{Y}|\times|\mathcal{Y}|}$ of the semantic information $\mathcal{A}$ as the class semantic similarities. Under the assumption that the inverse covariance $\Theta = \Sigma^{-1}$ is positive semidefinite, it minimizes an $\ell_{1}$-regularized negative log-likelihood:
\begin{equation}
\begin{aligned}
 \widehat{\Theta} = \underset{\Theta}{\argmin}~~\text{tr}(\mathcal{S}\Theta) - \log \text{det}(\Theta) + \delta\|\Theta\|_1,
\end{aligned}
\end{equation}
where $\mathcal{S}$ is a sample covariance matrix generated from $\mathcal{A}$, $\delta$ denotes the regularization parameter that controls the $\ell_1$ shrinkage. Thus, we have $\Sigma_{j,k}$ as the semantic similarity between classes $j$ and $k$, where $j,k \in [|\mathcal{Y}^{s,k}|]$.
We further take the semantic similarity matrix as the probability distribution that the prediction logits of a training sample should match with. A natural way of learning the probability distribution is through knowledge distillation with logits matching\cite{hinton2015distilling}. 

The class similarities $\Sigma$ are provided as the source knowledge to be transferred to target local models for learning visual features. We start with obtaining the soft targets by softening the peaky distribution of source and target logits with temperature scaling:
\begin{equation}
\begin{aligned}
&p_{j}^{\Sigma}(\Sigma_j;\tau) = \mathtt{softmax}(\Sigma_j; \tau) = \frac{\exp(\Sigma_j/ \tau)}{\sum_{m}\exp(\Sigma_m/\tau )},\\
&p_{j}^{a}(\widehat{\va}_{j}^{T} \mathcal{A};\tau) = \mathtt{softmax}(\widehat{\va}_{j}^{T} \mathcal{A}; \tau) = \frac{\exp(\widehat{\va}_{j}^{T} \mathcal{A}/ \tau)}{\sum_{m}\exp(\widehat{\va}_{m}^{T} \mathcal{A}/\tau )},
\end{aligned}
\end{equation}
where $\tau$ is the temperature that can produce a softer probability distribution over classes with a high value. We use $p^{\Sigma}_{j}$ and $p^{a}_{j}$ for short. The knowledge distillation loss measured by KL-divergence is:
\begin{equation}
\begin{aligned}
 \ell_{kl} &= \sum_{\vx_i \in \mathcal{D}^{s,k}} KL(p^{a}(\cdot; \tau) \| p^{\Sigma}(\cdot; \tau))\\
& =\tau^2 \sum_{\vx_i \in \mathcal{D}^{s,k}} \sum_{j}^{|\mathcal{Y}|}  p^{\Sigma}_{j}\log \frac{p^{\Sigma}_{j}}{p^a_{j}}.
\end{aligned}
\end{equation}

\noindent\textbf{Semantic-Visual Alignment.} To ensure the learned visual-semantic mapping is invertible, we further add a semantic-consistency loss to facilitate the visual feature reconstruction. In addition to the attribute regression layer $g(\cdot)$, we add another visual feature generation layer $h(\cdot)$ to construct visual features from the attributes $\widehat{\vv} = h(\va)\in\mathbb{R}^{d_v}$. The semantic-consistency loss is applied between the extracted visual features and the generated visual features:
\begin{equation}
\begin{aligned}
 \ell_{con} = \sum_{\vx_i \in \mathcal{D}^{s,k}} \| h(\va_{y_i}) - f(\vx_i) \|_2,
\end{aligned}
\end{equation}
where $\va_{y_i}$ is the corresponding class-level attributes for data sample $\vx_i$. 

\noindent \textbf{Joint Optimization.} The overall local objective function in the client $C_k$ for FedZSL is defined as:
\begin{equation}
\begin{aligned}
 \ell_{k} = \ell_{sce} + \mu\ell_{kl} + \ell_{con},
\end{aligned}
\end{equation}
where $\mu$ denotes the loss coefficient. The local models are trained with the overall local objective for a few epochs, the local update $\bigtriangleup w_t^k$ of the client $C_k$ in $t$-th communication round is submitted to the server for aggregation with Equation \ref{modelupdate}.%   

%-----------------------------------------------------------------------------------------
\begin{algorithm}[t]
% \hspace*{\algorithmicindent}
\hspace{-0.01em}\textbf{Input:} 
 clients number $K$, local datasets $\{\mathcal{D}^{s,1}, \mathcal{D}^{s,2}, \ldots, \mathcal{D}^{s,K}\}$, scaling factor $\beta$, communication round $T$, local epochs $E$, local learning rate $\lambda$, global learning rate  $\eta$, loss coefficient $\mu$, batch size $B$, FMD threshold $M$    \\
\hspace{-0.01em}\textbf{Initialize:} 
Server model parameters $w^{0}$
\caption{Federated Zero-Shot Learning (\texttt{FedZSL})}\label{euclid}
% \hrule
% \specialrule{.1em}{0.3em}{0.3em}
\begin{algorithmic}[1]
\State \textbf{Server executes:}
\For{$t = 0,1,...,T-1$}
    \State The server communicates $w^{t}$ to the \textit{i}-th client 
    \For{$k = 1,...,K$}
        \State The server communicates $w^{t}$ to the client $C_{k}$
        \State $\bigtriangleup w_t^k \leftarrow$ \textbf{PartyLocalTraining($k,w^t)$} 
    \EndFor
    \State $w_k^{t+1}$ $\leftarrow$  $w_t + \eta \frac{|\mathcal{Y}^{s,k}|\bigtriangleup w_t^k }{|\mathcal{Y}^{s}|}$
\EndFor

\State \textbf{PartyLocalTraining($k,w^t$):}
\State $w_{k}^t \leftarrow w^t$
\For{epoch $i = 0,1,\ldots,E$}
    \For{batch $\textbf{b} = \{\mX, \mY\} =   \{(\vx_b,y_b)\}^{B}$}
        \State $\mV \leftarrow f(\mX)$
        \For{$b = 0,1,\ldots,B$} 
            \State $\gamma_{b} = \sum_{c=1}^{d_v} \mV_{c,b}$
            \If{$\gamma_{b} < M$ }
            % \If{$\gamma_{b} = \sum_{c=1}^{d_v}f(\vx_b)< M$ }
                \State Discard $\mV_{b}$;~ $B$ -= $1$   % from the training batch
                % \State $B$ -= $1$ 
            \EndIf
        \EndFor
        \State $\{\widehat{a}\}^{B} \leftarrow g(\mV)$; $\{\widehat{v}\}^{B} \leftarrow \{h(\va_b\}\}^{B}$
        \State $\ell_k = \ell_{sce} + \mu \ell_{kd} + \ell_{con}$  by Eq. {\color{red} 3}, {\color{red} 6}, {\color{red} 7}
        \State $w_{k}^t \leftarrow w_{k}^t - \lambda \bigtriangledown \ell_k$ 
    \EndFor
\EndFor
\State $\bigtriangleup w_t^k \leftarrow \beta(w_{k}^t - w^t)$
\State Return $\bigtriangleup w_t^k$ to server
% \vspace{-1em}
\end{algorithmic}
% \hrule \\
% \hrulefill \\
% \specialrule{.01em}{0.00em}{0.3em}  \\
% \hspace{-0.01em} \textbf{Output: } 
\label{alg}
\end{algorithm}

\subsection{Feature Magnitude Backdoor Defense}
We consider backdoor attacks with model replacement \cite{bagdasaryan2020backdoor,bhagoji2019analyzing}, which is performed by a malicious client controlling a local device. The goal is to inject a backdoor functionality into the global model so that the specific inputs will be classified as a pre-defined class with high confidence. Without loss of generality, we assume the client 1 is malicious and attempts to replace the global model with a backdoored model $\widehat{w}$ with the updates:
\begin{equation}
\begin{aligned}
 \bigtriangleup w_t^1 = \beta (\widehat{w} - w_t),
\end{aligned}
\end{equation}
where $\beta$ is a scaling factor to ensure the backdoor functionality survives in server aggregation.

A conventional visual classifier produces discriminative visual features solely based on the semantic labels, so that these features are usually sparse in terms of semantics, \textit{i.e.,} various semantic information is covered. Instead, one of the properties in a zero-shot visual classifier is that the visual features are densely attended to the semantic attributes. Thus, we can regard the visual features as the entangled attribute information and the attribute regression layer is in charge of transforming or calculating the visual features' responses to each dimension in the attributes. As the dense visual features correspond to the annotated attributes, training samples with solid presences in the semantic attributes will show normal magnitude. In contrast, due to the absence of semantic information, as shown in Fig. \ref{fig:vis}, the visual features of malicious samples are usually sparse and of low magnitude. 

Based on the above insights, we construct a defending strategy, termed \textit{feature magnitude defense} (FMD). This strategy requires minimal changes to the FL protocol. By leveraging the magnitude mechanism of visual features, FMD can easily detect the malicious samples and remove them from training. Given a batch of input images 
$\mX = \{\vx_b\}^B $
, the CNN backbone extracts the visual features
: $\mV = f(\mX) \in \mathbb{R}^{d_v \times B}$
% : $\{\vv_i \}^{B} = \{f(\vx_i) \}^B$
, based on which the batch feature magnitude is defined as the sum: $\{\gamma_b\}^B = \{\sum^{d_v}_{c=1} \mV_{c,b}\}$. Within $\{\gamma_b\}^B$, if $\gamma_b$ is smaller than the magnitude threshold $M$, the corresponding training sample $\vx_b$ is conceived as malicious and removed from training. This defending procedure is presented in step 12 to 15 in Algorithm \ref{alg}.

%-------------------------------------------------------------------------
\section{Experiments}

\subsection{Setup}
\noindent \textbf{Datasets.} We extensively evaluate our method on several FL scenes with three zero-shot learning benchmark datasets. Caltech-UCSD Birds-200-20 (CUB) \cite{wah2011caltech} contains 11,700 images from 200 bird species with 312 manually annotated attributes. Animals with Attributes 2 (AwA2) \cite{xian2018zero} consists of 37,322 images from 50 animal classes with 85 attributes. SUN Scene Recognition (SUN) \cite{patterson2012sun} includes 14,340 images from 717 different scenes with 102 attributes. The standard splits for the seen and unseen classes, \textit{i.e.,} 150/50, 40/10, 645/72 are adopted on CUB, AwA2 and SUN respectively for the proposed split \cite{xian2018zero}.

\smallskip
\noindent \textbf{Evaluation Metrics.}
For the evaluation criteria, we use the average per-class top-1 accuracy as the primary metric in our conventional and generalized ZSL experiments (GZSL) \cite{xian2018zero}. 
In conventional ZSL setting, we only evaluate the accuracy of the \textit{unseen} classes $Acc_{\mathcal{C}}$ that none of the participants hold. In GZSL setting, we calculate the accuracies on test samples from both \textit{seen} and \textit{unseen} classes, \textit{i.e.,} $Acc_{\mathcal{Y}^{s}}$ and $Acc_{\mathcal{Y}^{u}}$. The harmonic mean $Acc_{\mathcal{H}}$ between the two sets of classes are computed to evaluate the GZSL performance: $ Acc_{\mathcal{H}} =  (2*Acc_{\mathcal{Y}^s}*Acc_{\mathcal{Y}^u})/(Acc_{\mathcal{Y}^s}+Acc_{\mathcal{Y}^u})$.

\smallskip
\noindent \textbf{Implementation Details.}
We use PyTorch to implement FedZSL and other baselines. Source code is made available in the supplementary material for reference. The local models are trained with an SGD optimizer with a weight decay of $1e-5$ and momentum of $0.9$. The learning rates for CUB, AwA2 and SUN are initialized as $5e-3, 3e-4$ and $1e-2$, respectively. The number of communication rounds and the default participant number are fixed to 50 and 10 unless explicitly specified. The batch size and number of local epochs in each communication round are set to 64 and 2. The coefficient $\mu$ of knowledge distillation is fixed to 3. For label smoothing, we set the temperature $\tau$ to 10. The $\ell_1$ penalty coefficient in Graphical Lasso is set to 0.01. A pre-trained lightweight CNN network ResNet18 is adopted as the backbone for all experiments.

\vspace{-5pt}
\subsection{Results and Analysis}
\vspace{-5pt}
\label{sec:exp}

\begin {table*}[t]
\caption {Performance comparisons (\%) on three datasets between various federated learning baselines and the proposed FedZSL. The best results of decentralized settings are highlighted in bold. }
\vspace{-10pt}
\begin{center}
\scalebox{0.85}{
\begin{tabular}[t]{lc ccc | cccc | cccc}
\toprule
      & \multicolumn{4}{c|}{CUB} & \multicolumn{4}{c|}{AwA2}  &  \multicolumn{4}{c}{SUN}   \\ 
  \cmidrule(lr){2-5} \cmidrule(lr){6-9} \cmidrule(lr){10-13} 

    & Acc$_{\mathcal{C}}$ &  Acc$_{\mathcal{Y}^{u}}$ & Acc$_{\mathcal{Y}^{s}}$ & Acc$_{\mathcal{H}}$  & Acc$_{\mathcal{C}}$ & Acc$_{\mathcal{Y}^{u}}$ & Acc$_{\mathcal{Y}^{s}}$ & Acc$_{\mathcal{H}}$  & Acc$_{\mathcal{C}}$ & Acc$_{\mathcal{Y}^{u}}$ & Acc$_{\mathcal{Y}^{s}}$ & Acc$_{\mathcal{H}}$ \\

  \noalign{\smallskip}
 \hline 
\noalign{\smallskip}
% \multirow{6}{*}{\dag}   
 \texttt{Baseline($\varphi=0.1$)}        
    & 14.9          & 8.2               & 7.4             & 7.7
    & 29.5          & 16.5              & 12.6            & 14.3
    & 25.6          & 11.1              & 5.5             & 7.4
    \\

% \multirow{10}{*}{\ddag} 
 \texttt{Baseline($\varphi=0.2$)}     
    & 27.0          & 16.1              & 13.5            & 14.7      
    & 32.6          & 22.6              & 17.3            & 19.6   
    & 34.9          & 16.9              & 8.3             & 11.1    
    \\

% \multirow{10}{*}{\ddag} 
 \texttt{Baseline($\varphi=0.3$)}     
    & 32.1          & 22.6              & 19.0            & 20.6 
    & 43.3          & 25.0              & 30.2            & 27.3   
    & 38.6          & 18.0              & 10.4            & 13.2  
 
    \\
    
     \texttt{Baseline($\varphi=0.5$)}     
    & 43.9          & 30.5              & 26.5            & 28.3      
    & 45.7          & 26.0              & 36.0            & 30.2   
    & 41.3          & 25.8              & 14.4            & 18.5   
    \\
\noalign{\smallskip}
 \hline 
\noalign{\smallskip}
% \multirow{10}{*}{\ddag} 
 \texttt{FedAvg}  
    & 44.6          & 30.3              & 30.7            & 30.5      
    & 39.5          & 27.5              & 34.5            & 31.0   
    & 42.4          & 25.4              & 12.2            & 16.5    
    \\
\texttt{FedAvg-decay}
& {45.8}          & {31.5}              & {32.1}            & {31.8}
& {40.0}          & {26.8}              & {35.2}            & {30.4}
& {43.1}          & {26.8}              & {13.1}            & {17.6}
\\
   \texttt{FedProx}
& {44.3}          & {29.5}              & {31.8}            & {30.6}
& {39.7}          & {26.2}              & {34.3}            & {29.7}
& {42.0}          & {25.1}              & {11.8}            & {16.1}
\\
% \cmidrule[.1em]{2-14}
% \texttt{FedZSL$_{\text{Imbalanced}}$}
%     & 51.9          & 36.9              & 39.4           & 38.1      
%     & 46.3          & 36.7              & \textbf{52.1}           & 43.1   
%     & 47.4          & 24.6              & 15.3           & 18.8    
%     \\
\noalign{\smallskip}
 \hline 
\noalign{\smallskip}
  \texttt{FedZSL}(ours)      
& {55.1} &{ 43.6} & {40.3} & {41.9}
& {46.6 }         & {37.8 }             & {50.8}            & {43.4}
& {49.2 }         & {27.6}              & \textbf{16.8 }           & {20.9}
\\

  ours+\texttt{FedAvg-decay}
& \textbf{56.3}   & \textbf{43.8}       & \textbf{40.4}     & \textbf{42.0}
& \textbf{47.2}          & {37.9}              & \textbf{51.3}            & \textbf{43.6}
& \textbf{49.6}          \textbf& {28.9}              & {16.7}            & \textbf{21.2}
\\

  ours+\texttt{FedProx}      
& {55.6}          & {42.9}              & {39.7}            & {41.2}
& {45.9}          & \textbf{38.1}              & {50.2}            & {43.3}
& {48.6}          & {27.3}              & {16.4}            & {20.5}
\\
\noalign{\smallskip}
 \hline 
\noalign{\smallskip}
\texttt{Centralized}     
    & 55.4          & 42.7              & 44.9            & 43.8      
    & 59.0          & 35.0              & 72.6            & 47.2   
    & 50.1          & 26.7              & 20.2            & 23.0    
    \\
    \noalign{\smallskip}
\hline\bottomrule
\end{tabular}}
\end{center}
% \vspace{-30pt}
\label{gzslperoformance}
\end {table*}

Table \ref{gzslperoformance} reports the performance comparisons among several federated learning models for zero-shot recognition.

\begin{figure}
    \centering
    \includegraphics[width=1\linewidth]{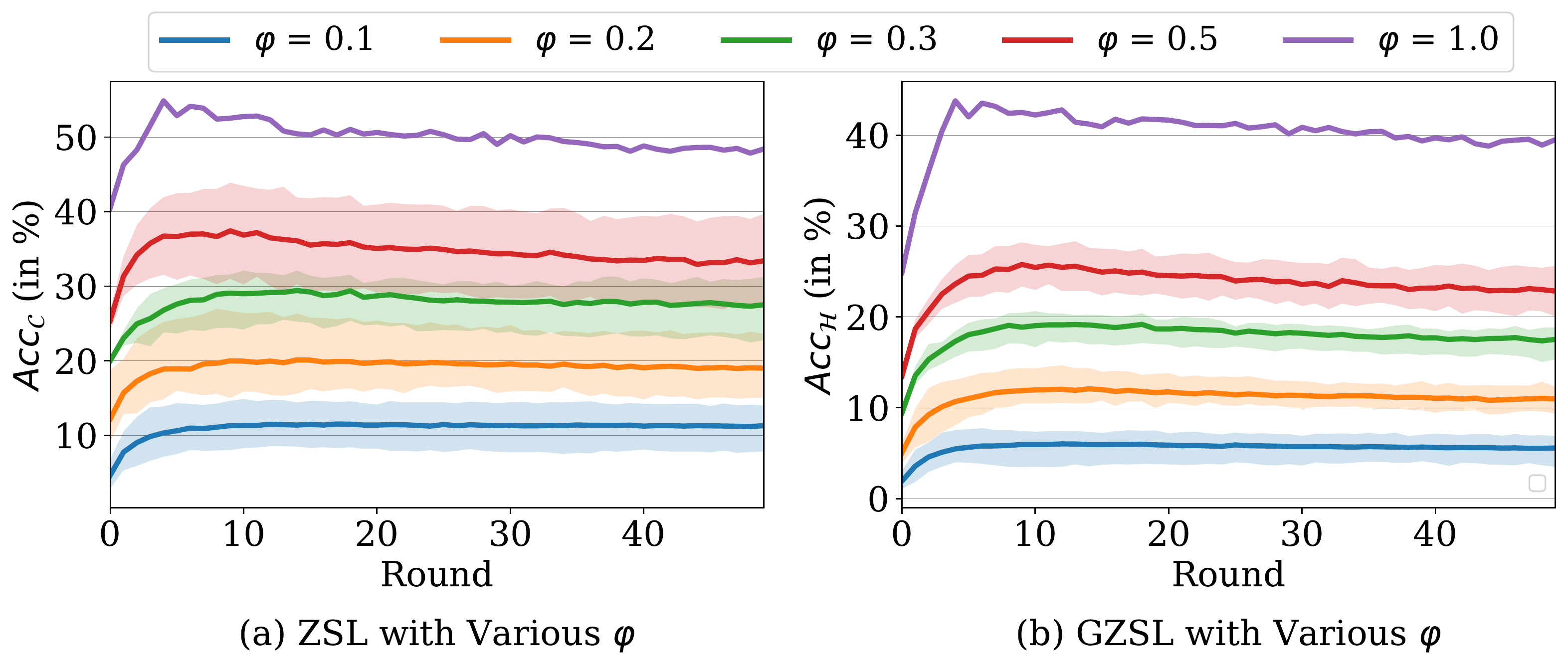}
    \vspace{-15pt}
    \caption{Learning curves of decentralized baselines for the CUB dataset in the ZSL and GZSL settings. The solid line indicates the averaged performance, and the shaded area represents the performance variance across local devices.}
    \label{fig:differentclasses}
    % \vspace{-15pt}
\end{figure}

\noindent \textbf{Baselines and Centralized:} Specifically, we decentralize the data onto 10 clients, with each client only having a subset of classes. 
For simplicity, we define the local class ratio $\varphi = \{0.1, 0.2, 0.3, 0.5\}$ to indicate the percentage of the total seen classes available in each device. For $\varphi=1$, it degenerates to the centralized training (\texttt{Centralized}),  which leverages all the seen classes to train a zero-shot classifier. From rows 1-4, it is observed that when training with more seen classes,  the model can  generalize better to unseen classes,  which is in line with our assumption. Fig.\ref{fig:differentclasses} depicts the detailed learning curves of baselines with various portions of the seen classes available in each round. With models trained separately in local devices, it demonstrates a significant variance (the shaded area) of model performance across devices. From the solid line, we can see that more seen classes generally lead to a better performance in both ZSL and GZSL.

\smallskip

\noindent \textbf{Federated Baselines:} We compare the proposed approach with the most representative federated learning frameworks \texttt{FedAvg} \cite{mcmahan2017communication}, \texttt{FedAvg-decay}  \cite{li2019convergence}, and \texttt{FedProx} \cite{li2020federated}. FedAvg learns a global model by aggregating parameters from local devices, which is nearly the same as \texttt{Baseline ($\varphi=0.1$)} but with an additional server aggregation step. Table \ref{gzslperoformance} shows that \texttt{FedAvg} with $\varphi=0.1$ can obtain comparable results with baseline ($\varphi=0.5$). As clients do not take advantage of the semantic attributes, FedAvg fails to achieve significant performance gains for the unseen classes over baselines. FedAvg-decay has an extra requirement on the global learning rate. By decaying the learning rate of global aggregation, the performance results on CUB and SUN are improved comparing with FedAvg. FedProx is a federated baseline tackling heterogeneity, which assigns higher aggregation rates to the clients that hold more training data. However, our experiments show that in our proposed \textit{p.c.c.d.} setting, simply associating aggregation rate with the amount of data fails to achieve better performance.

\smallskip

\noindent \textbf{FedZSL:} With the proposed CARD training strategy that helps reach the consensus across devices, the proposed \texttt{FedZSL} outperforms all decentralized baselines and \texttt{FedAvg} in both ZSL and GZSL scenarios.  When applying the proposed CARD training strategy on \texttt{FedAvg-decay}, in ours+\texttt{FedAvg-decay}, the global learning rate decay can slightly improve the performance. In the variant ours+\texttt{FedProx}, as the clients hold the same number of training classes, assigning different global aggregation rate causes imbalanced knowledge transfer across devices, leading to inferior global performance.

\smallskip
\begin{figure}
    \centering
    \vspace{-8.5pt}
    \includegraphics[width=1\linewidth]{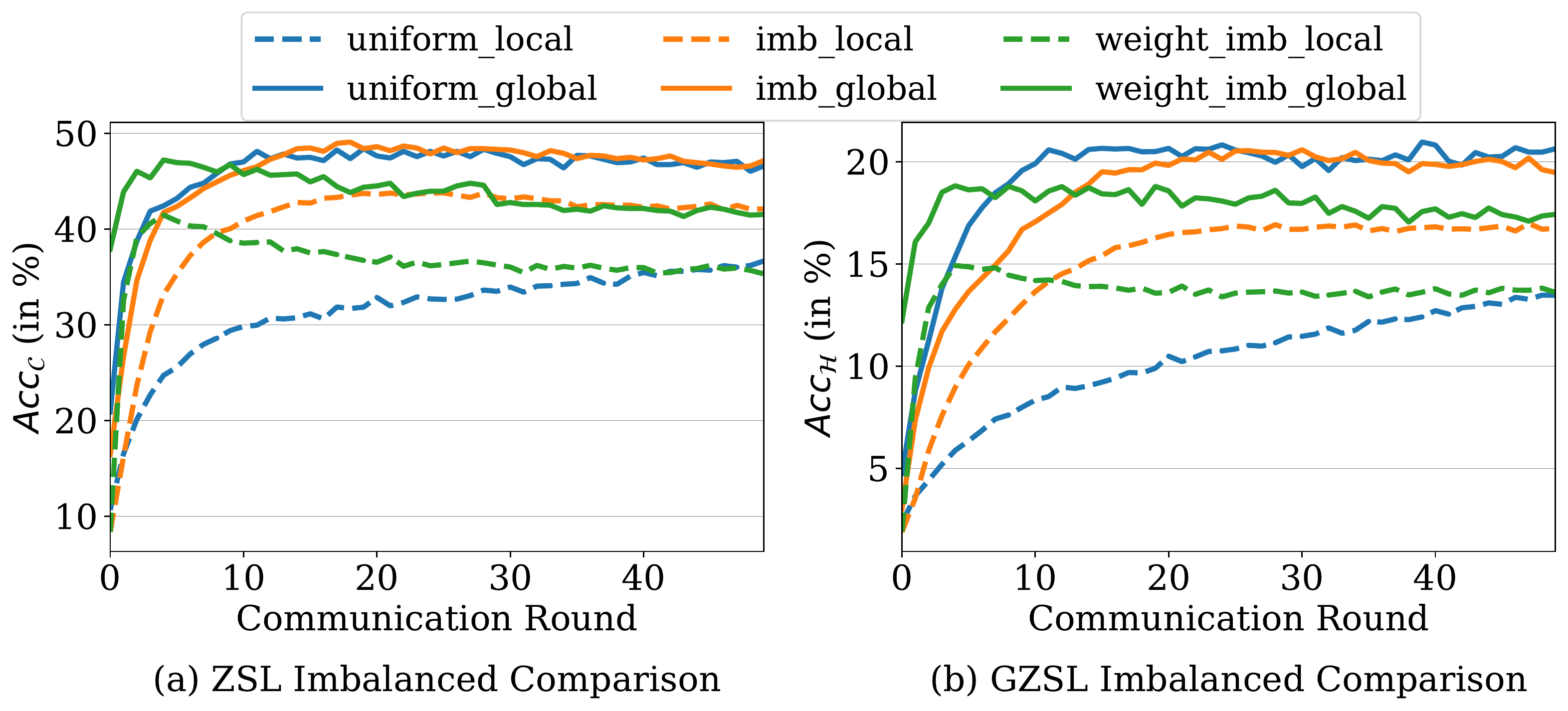}
    \vspace{-16pt}
    \caption{Performance comparisons of the proposed FedZSL with (1) \textit{uniform}: uniform class distribution, (2) \textit{imb}: imbalanced class distribution on local devices. \textit{weight} plots indicate the performance with weighted aggregation instead of averaging.}    % Local: average performance of the local models, global: performance of the aggregated model. 
    \label{fig:imbalanceperformance}
    % \vspace{-16pt}
\end{figure}

\noindent \textbf{Impact of Imbalanced Class Distribution.} Considering participants are not always holding the identical number of classes in real-world cases, we conduct experiments in a more practical setting with the SUN dataset, which is to learn a global model from imbalanced class distribution. The statistics of class distribution can be found in the Appendix. In Fig. \ref{fig:imbalanceperformance},  we compare the training convergence among local and global models trained with (1) uniform class distribution and (2) imbalanced class distribution. For class imbalance, we further make a comparison of average aggregation (in orange) and weighted aggregation (in green) based on the ratio of available classes. It is shown that even with imbalanced class distribution, the global model can reach a similar performance to the one trained with the uniform class distribution. An interesting observation is that when weighting the local model updates according to the local class number, the global model can converge quickly, but the performance is inferior to average weighting.

\begin{figure*}[t]
    \centering
    \includegraphics[width=1\linewidth]{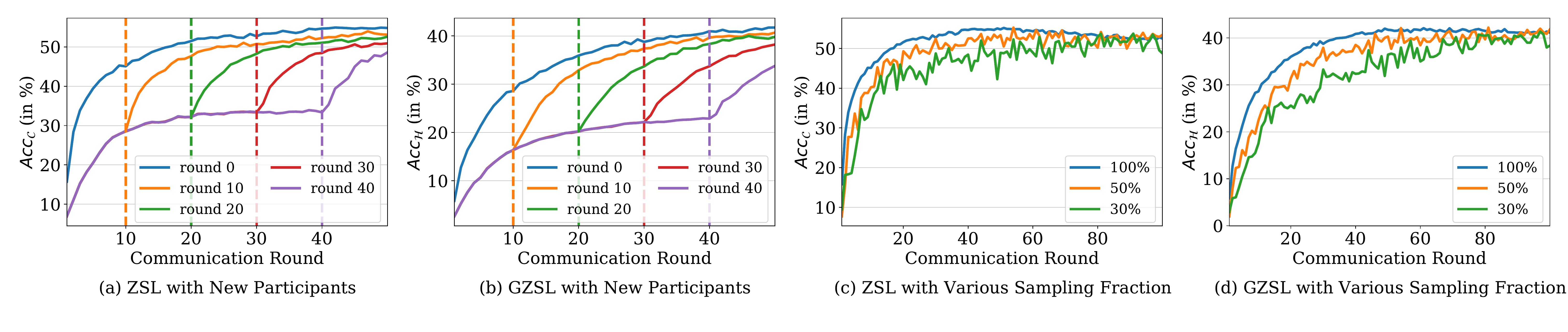}
    % \vspace{-15pt}
    \caption{(a) and (b): ZSL and GZSL learning curves with more participants joining in round 10, 20, 30, 40. (c) and (d): ZSL and GZSL learning curves with various sampling fractions. }
    \label{fig:moreclasses}
    % \vspace{-10pt}
\end{figure*}

\smallskip
\noindent \textbf{Impact of New Participants.}
To investigate the performance influence when new seen classes participate in the training, we conduct experiments by adding more seen classes when the federated model is in different stages. In Fig. \ref{fig:moreclasses}(\textcolor{red}{a}) and \ref{fig:moreclasses}(\textcolor{red}{b}), we start by training with only 5 participants, each of which contains 15 seen classes. Then, we add another 5 participants, each with 15 seen classes from round 0, 10, 20, 30, and 40. We can see that the performance can quickly surge to the performance level as training with all participants. This result indicates that \texttt{FedZSL} can quickly adapt to new seen classes, and the emerging of new participants does not cause a negative impact on training.

\smallskip
\noindent \textbf{Impact of Various Sampling Fractions.}
To investigate the impact of sampling fraction, we conduct experiments by choosing 30\%, 50\%, and 100\% of the participants in each communication round to observe the convergence rate. It can be seen from Fig. \ref{fig:moreclasses}(\textcolor{red}{c}) and \ref{fig:moreclasses}(\textcolor{red}{d}), after 100 communication rounds, the performance results on both ZSL and GZSL reach a similar level. In addition, sampling fewer participants results in a slower convergence rate and unstable performance when aggregating the local models trained on partial seen classes.

\begin {table*}[t]
\caption {Effects of different components on three datasets with various stripped-down versions of our full proposed model.}
\vspace{-2pt}
\begin{center}
\scalebox{0.85}{
\begin{tabular}[t]{lc ccc | cccc | cccc}
\toprule
      & \multicolumn{4}{c|}{CUB} & \multicolumn{4}{c|}{AwA2}  &  \multicolumn{4}{c}{SUN}   \\ 
  \cmidrule(lr){2-5} \cmidrule(lr){6-9} \cmidrule(lr){10-13} 

    & Acc$_{\mathcal{C}}$ &  Acc$_{\mathcal{Y}^{u}}$ & Acc$_{\mathcal{Y}^{s}}$ & Acc$_{\mathcal{H}}$  
    & Acc$_{\mathcal{C}}$ & Acc$_{\mathcal{Y}^{u}}$ & Acc$_{\mathcal{Y}^{s}}$ & Acc$_{\mathcal{H}}$  
    & Acc$_{\mathcal{C}}$ & Acc$_{\mathcal{Y}^{u}}$ & Acc$_{\mathcal{Y}^{s}}$ & Acc$_{\mathcal{H}}$ \\

  \noalign{\smallskip}
 \hline 
\noalign{\smallskip}
 \texttt{FedAvg}  
    & 44.6          & 30.3              & 30.7            & 30.5      
    & 39.5          & 27.5              & 34.5            & 31.0   
    & 42.4          & 25.4              & 12.2            & 16.5    
    \\

  {\texttt{FedZSL} w/o $\ell_{kl}$}      
& {48.1} &{ 38.4} & {39.6} & {39.0}
& {43.2}         & {32.5 }             & {47.9}            & {38.7}
& {43.8}         & {27.5}              & {14.9}           & {19.3}
\\
  {\texttt{FedZSL} w/o $\ell_{con}$ }      
& {54.0} &{39.7} & {41.2} & {40.4}
& {45.8}         & {35.6}              & {49.1}            & {41.3}
& {47.9}         & {26.8}              & {16.1}           & {20.1}
\\
  {\texttt{FedZSL}(ours)}      
& \textbf{55.1} &\textbf{ 43.6} & \textbf{40.3} & \textbf{41.9}
& \textbf{46.6 }         & \textbf{37.8 }             & \textbf{50.8}            & \textbf{43.4}
& \textbf{49.2 }         & \textbf{27.6}              & \textbf{16.8 }           & \textbf{20.9}
\\
\noalign{\smallskip}
\hline\bottomrule
\end{tabular}}
\end{center}
\vspace{-10pt}
\label{ablation}
\end {table*} 

\smallskip

\noindent \textbf{Ablation Study.}
In this ablation study, we evaluate various stripped-down versions of our full proposed model to
validate different components of the proposed FedZSL. In Table \ref{ablation}, we report the ZSL and GZSL performance of each version on the three benchmark datasets. The best performance is achieved when the CARD training strategy with the KL loss function $\ell_{kl}$ and the semantic-visual alignment with loss $\ell_{con}$ are both applied.

\subsection{Backdoor Attacks} 
\begin{figure}[t]
    \centering
    \includegraphics[width=1\linewidth]{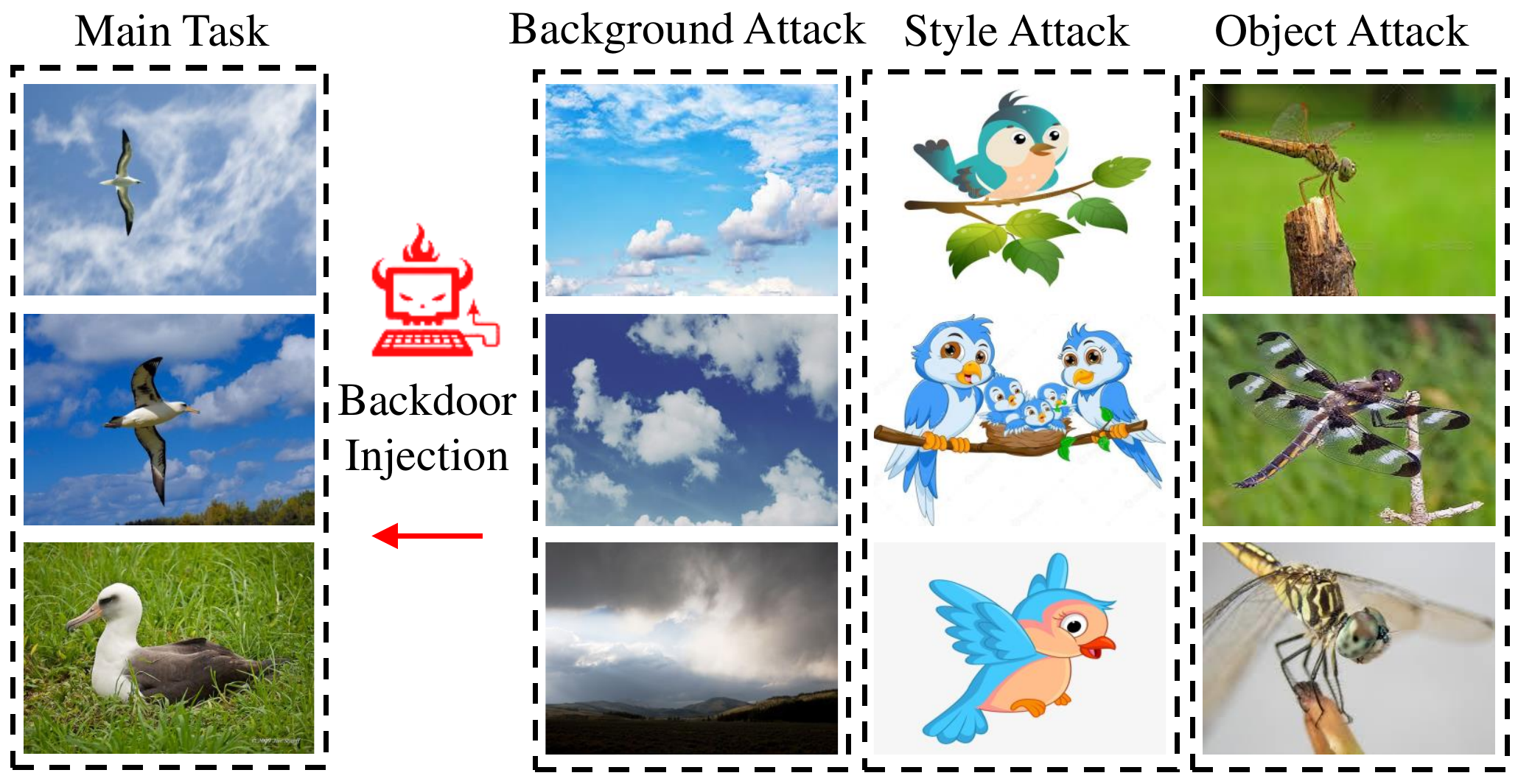}
    % \vspace{-10pt}
    \caption{Exemplars of three different types of backdoor attacks.}
    \label{fig:backdoorsamples}
    \vspace{15pt}
\end{figure}

To investigate the impact of backdoor attacks on \texttt{FedZSL}, we perform a series of experiments on how to backdoor and how to defend backdoors in \texttt{FedZSL} with the proposed feature magnitude mechanism. There are three different backdoors we consider in this paper, including background attack, style attack, similar object attack.  Correspondingly, we choose sky, cartoon birds, and dragonflies as malicious samples with 50 images per class. Fig. \ref{fig:backdoorsamples} shows some samples of three different types of backdoor attacks. The experiments are mainly conducted on the CUB dataset.

\begin{figure}
    \centering
    \includegraphics[width=1\linewidth]{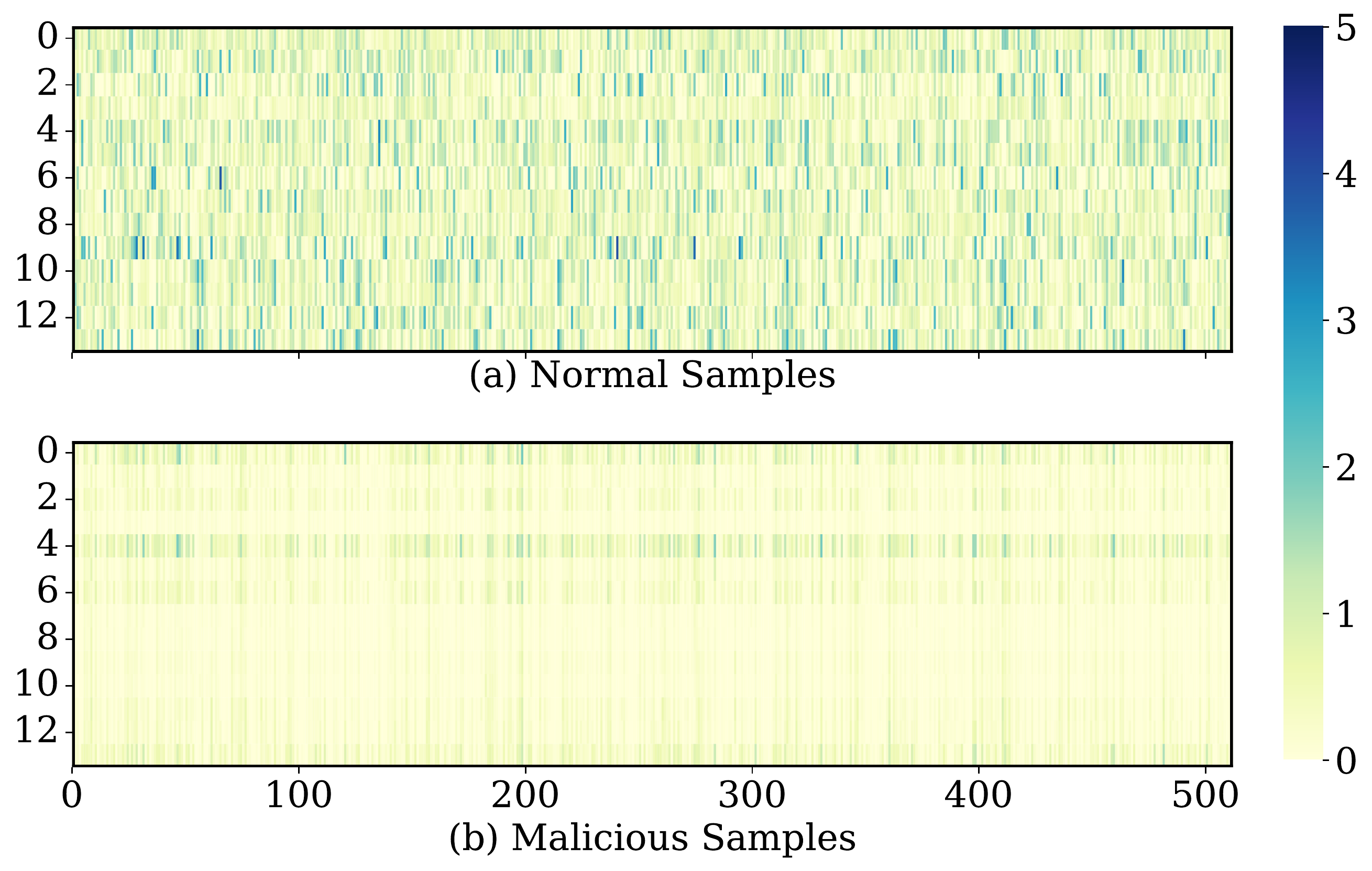}
    % \vspace{-15pt}
    \caption{Visualization of feature magnitude for normal and malicious samples.}
    \label{fig:vis}
    % \vspace{-10pt}
\end{figure}

\smallskip
\noindent \textbf{Feature Magnitude Visualization.}
In Fig. \ref{fig:vis}, we visualize the visual features of normal and malicious samples. It can be seen that in contrast to the normal samples, the visual features of malicious samples are of low magnitude. As we state in the methodology section, the visual features produced by a zero-shot learning classifier are essentially the presences of the semantic information. Thus, the zero-shot learning classifier could not find the visual representations corresponding to the annotated attributes for a malicious sample, resulting in visual features with low magnitude.

\begin{figure*}
    \centering
    \includegraphics[width=1\linewidth]{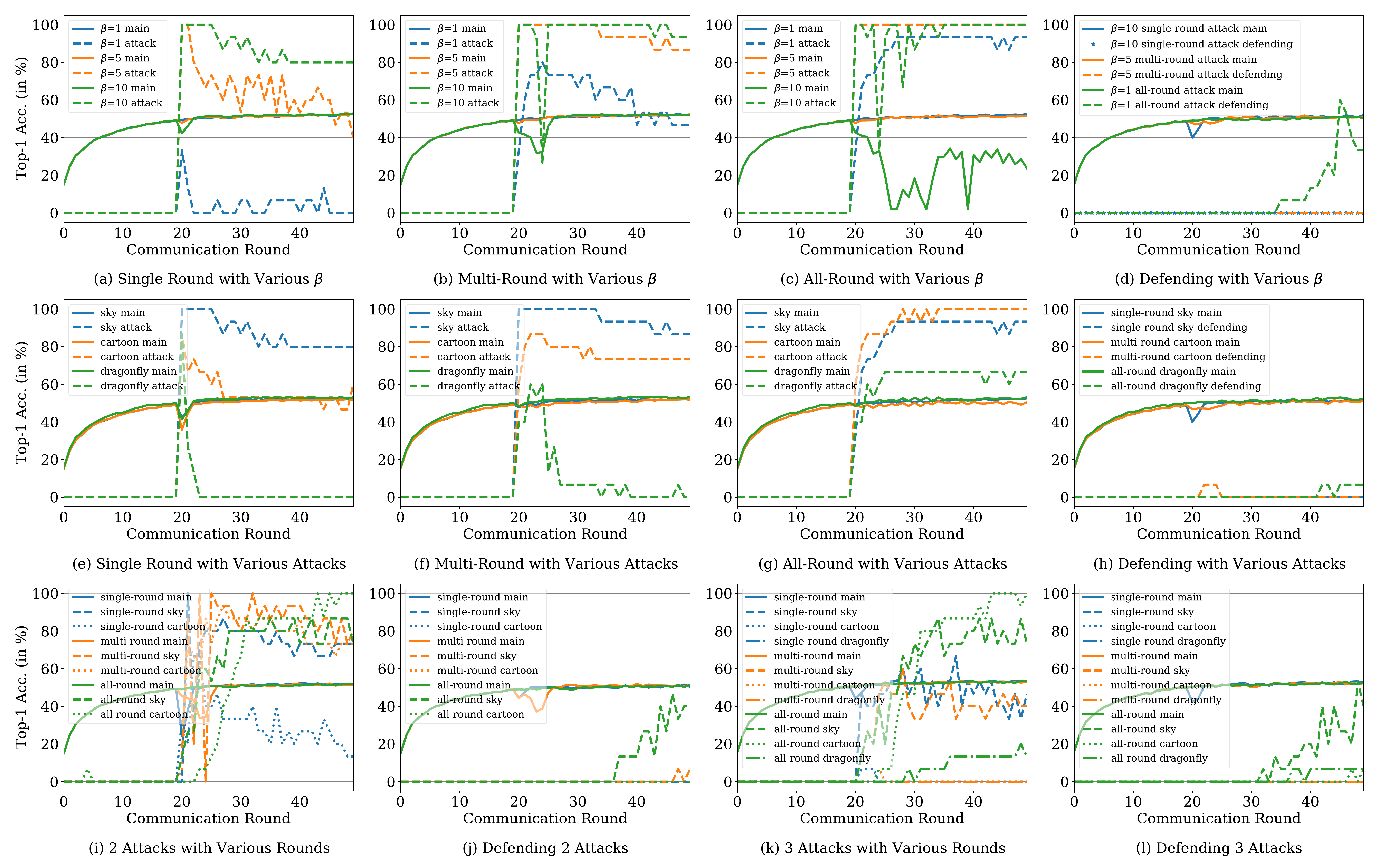}
    \vspace{-20pt}
    \caption{Learning curves of various attacks and corresponding defending with feature magnitude estimation.}
    \label{fig:attack}
    \vspace{-10pt}
\end{figure*}

\smallskip
\noindent \textbf{Impact of the Scaling Factor.}
In Fig. \ref{fig:attack}(\textcolor{red}{a})-\ref{fig:attack}(\textcolor{red}{d}), we study the impact of the scaling factor on the backdoor accuracy. Specifically, in Fig. \ref{fig:attack}(\textcolor{red}{a}), we test with scaling factor $\beta$ in $\{1,5,10\}$ in single-round attack (round 20), it turns out that the backdoor functionality could be canceled out soon after a few rounds of server aggregation with a small scaling factor. When setting $\beta$ to 10, the backdoor accuracy remains high after 30 rounds of normal aggregation. In Fig. \ref{fig:attack}(\textcolor{red}{b}), backdoor samples are injected from rounds 20 to 25. We can see that when we set $\beta$ to 10, the main task performance drops significantly and recover to the normal standard after a few rounds. 
In Fig. \ref{fig:attack}(\textcolor{red}{c}), we inject attack samples in all rounds after 20, which shows when setting $\beta$ to 1, the backdoor accuracy can keep high and main task performance is not affected. In Fig. \ref{fig:attack}(\textcolor{red}{d}), our proposed FMD is used to detect and remove the malicious sample from training, and we can completely defend the backdoor attacks for single- and multi-round attacks. For the all-round attacks, we can substantially reduce the backdoor accuracies. 

\smallskip
\noindent \textbf{Impact of Attack Types.}
To investigate the attack difficulty of different attack types, in Fig. \ref{fig:attack}(\textcolor{red}{e})-(\textcolor{red}{g}), we can see that the background attack is the easiest one to inject. The style attack is also considerably easy to inject, and through more communication rounds, the accuracy is increased. As for the similar object, only all-round attack can successfully inject the backdoor functionality. In Fig. \ref{fig:attack}(\textcolor{red}{h}), we show that our defending method works for all types of attacks. 

\smallskip
\noindent \textbf{Impact of Simultaneous Attacks.}
In Fig. \ref{fig:attack}(\textcolor{red}{i})-(\textcolor{red}{l}), we test backdoor attacks with multiple compromised participants. In Fig. \ref{fig:attack}(\textcolor{red}{i}), both background and style attacks are injected. It can be seen that even though a single-round attack, the two backdoor accuracies remain considerably high after many communication rounds. With more attack rounds in multi-round and all-round attacks, the attack accuracies increase as expected. We show that our defending method consistently prevents  malicious injections in Fig. \ref{fig:attack}(\textcolor{red}{j}). A more extreme experiment is conducted to inject all three kinds of attacks at the same time. In Fig. \ref{fig:attack}(\textcolor{red}{k}) we can see that it is very hard to inject all three backdoors at the same time successfully, and by introducing FMD in Fig. \ref{fig:attack}(\textcolor{red}{l}) we can substantially reduce all attack accuracies.

%-------------------------------------------------------------------------

\begin{figure}
    \centering
    \vspace{-10pt}
    \includegraphics[width=1\linewidth]{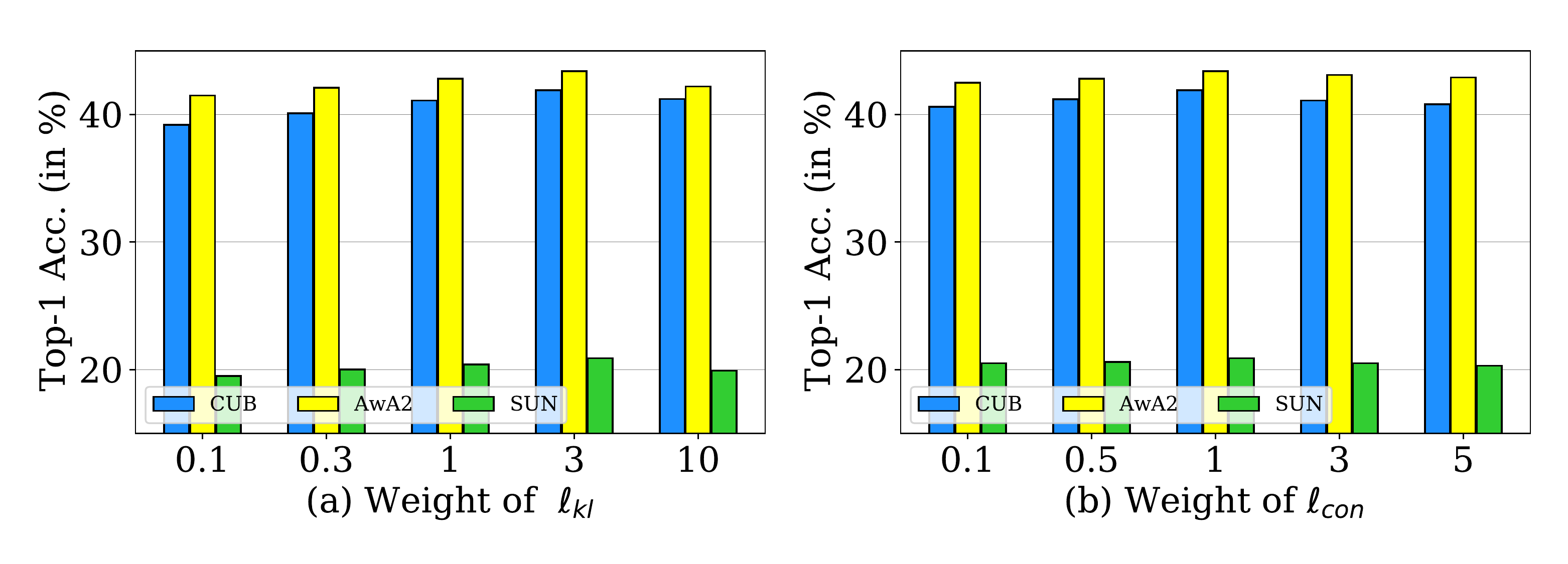}
    % \vspace{-25pt}
    \caption{ Hyper-parameter sensitivity for $\ell_{kl}$ and $\ell_{con}$. }
    \label{fig:hyper}
    % \vspace{-20pt}
\end{figure}

% \begin{figure}
%     \centering
%     \includegraphics[width=85mm]{figures/attack_defend.pdf}
%     \caption{(a) Defending single round (round 20) single attack with different scaling factor $\beta$. (b) Defending multi-round (round 20-24) single attack with different scaling factor $\beta$}.
%     \label{fig:attack2}
%     \vspace{-10pt}
% \end{figure}
\noindent \textbf{Hyper-parameter Sensitivity.}
There are mainly two hyper-parameters controlling the overall objective function for the local model training, including the weight of $\ell_{kl}$ and $\ell_{con}$. To better understand the effects of the different components in the proposed \texttt{FedZSL}, we report the sensitivity visualization of the two hyper-parameters in Fig. \ref{fig:hyper}. It can be seen that the performance of all three datasets peaks when the weight of $\ell_{kl}$ is set to 3 and $\ell_{con}$ is set to 1.

\begin{figure}
    \centering
    % \vspace{-15pt}
    \includegraphics[width=0.9\linewidth]{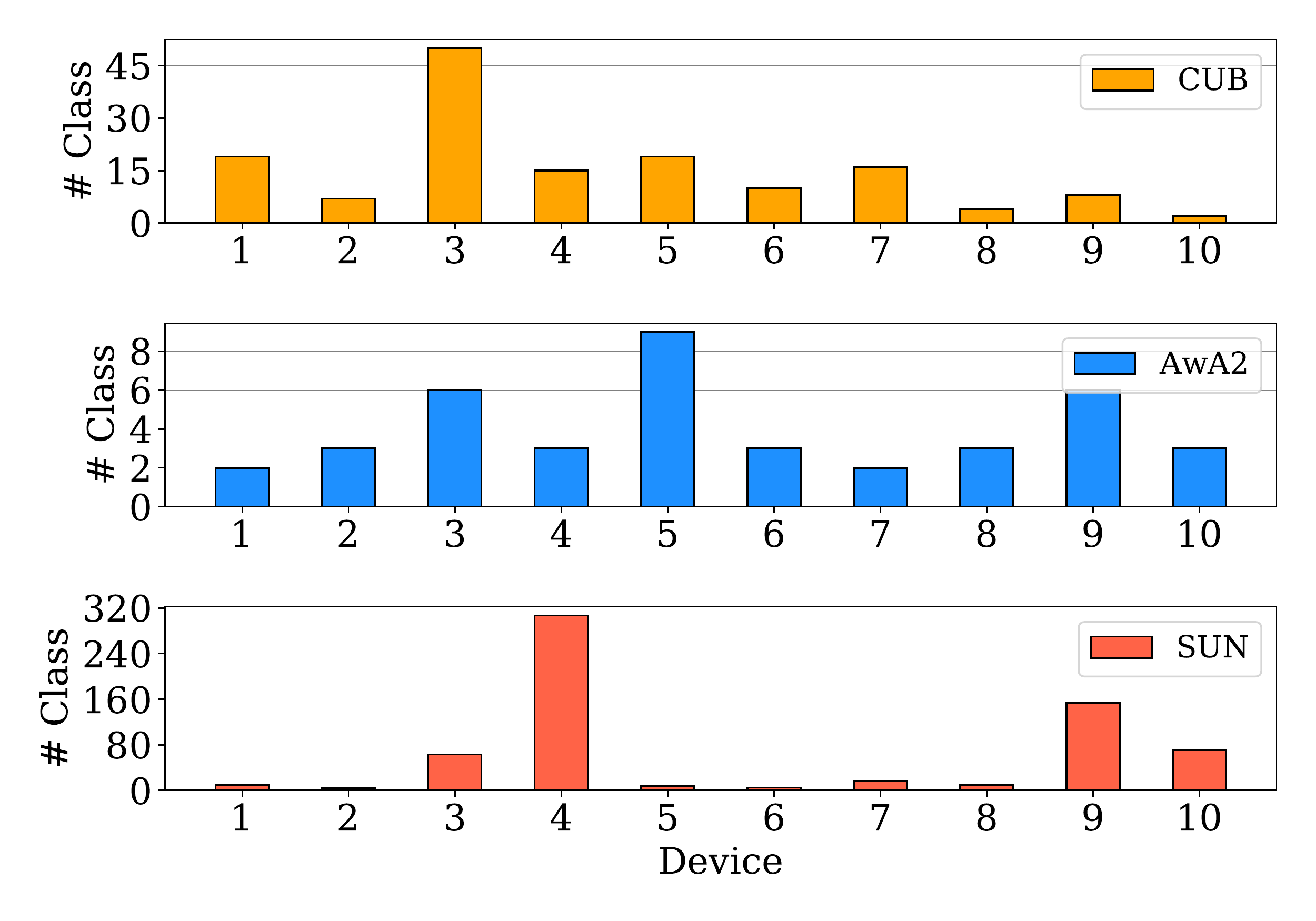}
    \vspace{-16pt}
    \caption{Imbalanced class distributions on three zero-shot datasets.}
    \label{fig:imbalanced}
    % \vspace{-25pt}
\end{figure}

\smallskip
\noindent \textbf{Generation of Imbalanced Class Distribution.} 
In Section \textcolor{red}{4.2}, we discussed the impact of imbalanced class distribution on learning curves in contrast to the uniform class distribution. To help understand what the imbalanced class distribution looks like, in Fig. \ref{fig:imbalanced} we visualize the number of classes that each device possesses. It can be seen that there is a large variance of class numbers across %
devices. As for implementation, the imbalanced class partition is sampled with the Dirichlet distribution. Specifically, we sample $\vp \sim Dir(\alpha)$ and $\vp_k$ is the proportion of the classes to be allocated to the client $C_k$. $Dir(\alpha)$ is the Dirichlet distribution with a concentration parameter $\alpha$ (0.5 by default). The higher the $\alpha$, the more uniform the distribution.  Another distribution generation constraint is that we set the minimum class number on each device to 2.

\begin{figure*}[t]
  \begin{minipage}[t]{0.35\linewidth}
    \includegraphics[width=\linewidth]{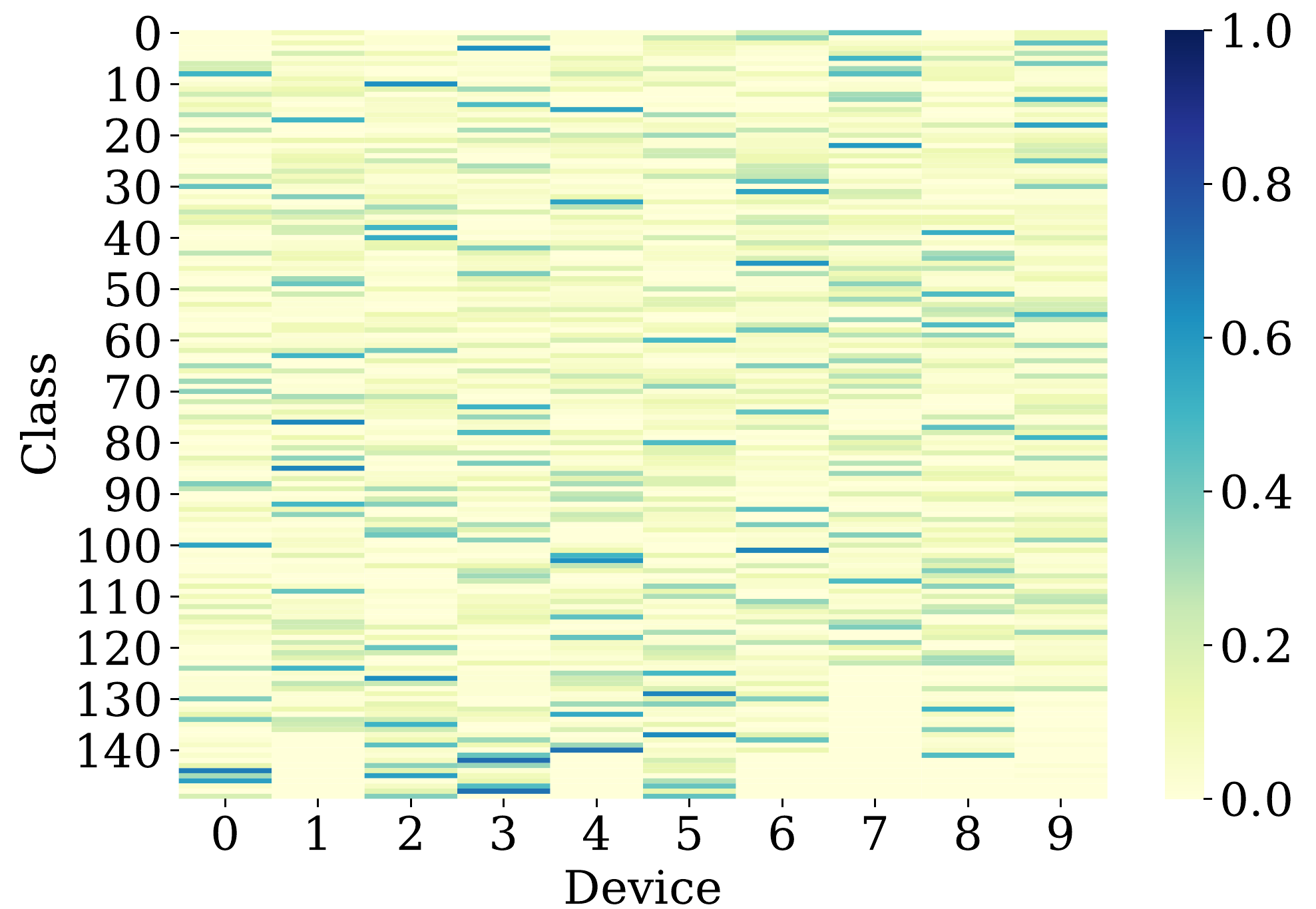}
    \caption{The data distribution of each device using non-\textit{i.i.d.} data partition.} 
    % The color bar denotes the proportion of data samples in the corresponding class.}
    \label{fig:noiid}
  \end{minipage}\hfill%
  \begin{minipage}[t]{0.60\linewidth}
    \includegraphics[width=\linewidth]{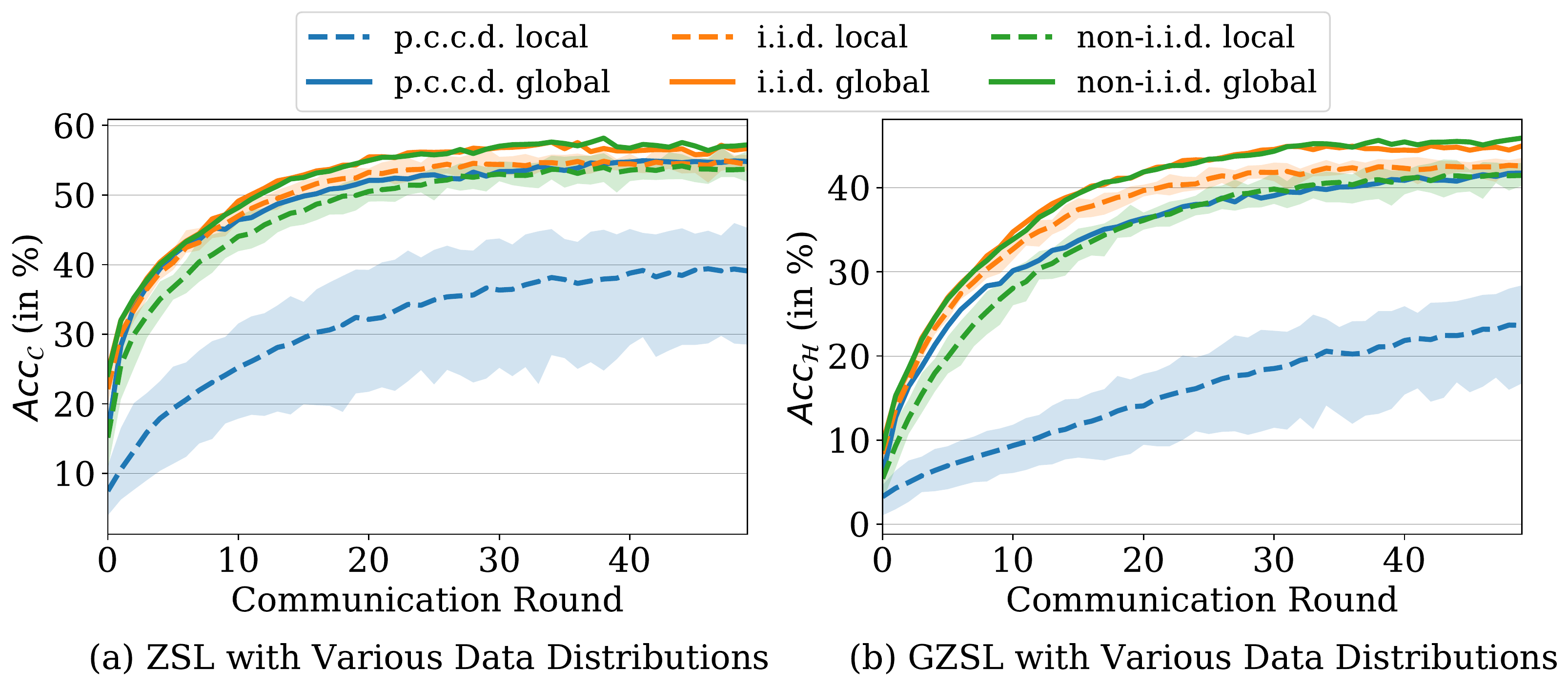}
    \caption{ZSL and GZSL learning curves with different data distributions (\textit{i.e.,} \textit{i.i.d.}, non-\textit{i.i.d.}, and our proposed \textit{p.c.c.d.})}
    \label{fig:iid}
  \end{minipage}
\end{figure*}

\smallskip
\noindent \textbf{Impact of \textit{i.i.d.}, non-\textit{i.i.d.} and \textit{p.c.c.d.} Distributions.}
% Our motivation is to learn a zero-shot model with \textit{p.c.c.d} data distribution across devices.
To investigate the impact of different data distributions, we also conducted experiments with the data sampled from \textit{i.i.d.} and non-\textit{i.i.d.} distributions to compare with the model trained in our \textit{p.c.c.d.} setting. In \textit{i.i.d.}, we uniformly split the set of the seen classes onto ten devices, whereas in non-\textit{i.i.d.} we use a Dirichlet distribution $Dir(\alpha)$ with a concentration parameter $\alpha=0.5$ to sample the data partition.
The non-\textit{i.i.d.} data partition across devices is visualized in Fig. \ref{fig:noiid}, in which each device has relatively few (even 0) training samples in some classes. In Fig. \ref{fig:iid}, we illustrate the learning curves with three different types of data distributions. The dashed lines and shaded areas represent the performance and its variance across local devices before the global aggregation in each communication round;

% \begin{figure}
%     \centering
%     % \vspace{-8pt}
%     \includegraphics[width=0.7\linewidth]{figures/noniid.pdf}
%     % \vspace{2pt}
%     \caption{The data distribution of each device using non-\textit{i.i.d.} data partition. The color bar denotes the proportion of data samples in the corresponding class.}
%     \label{fig:noiid}
%     % \vspace{-25pt}
% \end{figure}

% \begin{figure}
%     \centering
%     % \vspace{-14pt}
%     \includegraphics[width=0.95\linewidth]{figures/iid.pdf}
%     \caption{ZSL and GZSL learning curves with different data distributions (\textit{i.e.,} \textit{i.i.d.}, non-\textit{i.i.d.}, and our proposed \textit{p.c.c.d.})}
%     \label{fig:iid}
%     % \vspace{-25pt}
% \end{figure}
% The shaded area is the performance variance of local models across devices. 

\noindent the solid lines indicate the global performance for different distributions. 
As the devices in our \textit{p.c.c.d.} are trained on various classes, the performance variance as indicated in the shaded area is more significant. Fig. \ref{fig:iid} also confirms that the %
\noindent non-\textit{i.i.d.} setting witnesses a slightly higher variance than the \textit{i.i.d.} setting, which is resulted from that the sample numbers of different classes vary across clients in the non-\textit{i.i.d.} protocol. For the global performance, the recognition accuracy of the proposed model trained in the challenging \textit{p.c.c.d.} setting is only slightly inferior to ones trained in the \textit{i.i.d.} and non-\textit{i.i.d.} settings. This further verifies the robustness of the proposed FedZSL framework, which is agnostic to the different data distributions on local devices.

\begin{figure}
    \centering
    % \vspace{-12pt}
    \includegraphics[width=1\linewidth]{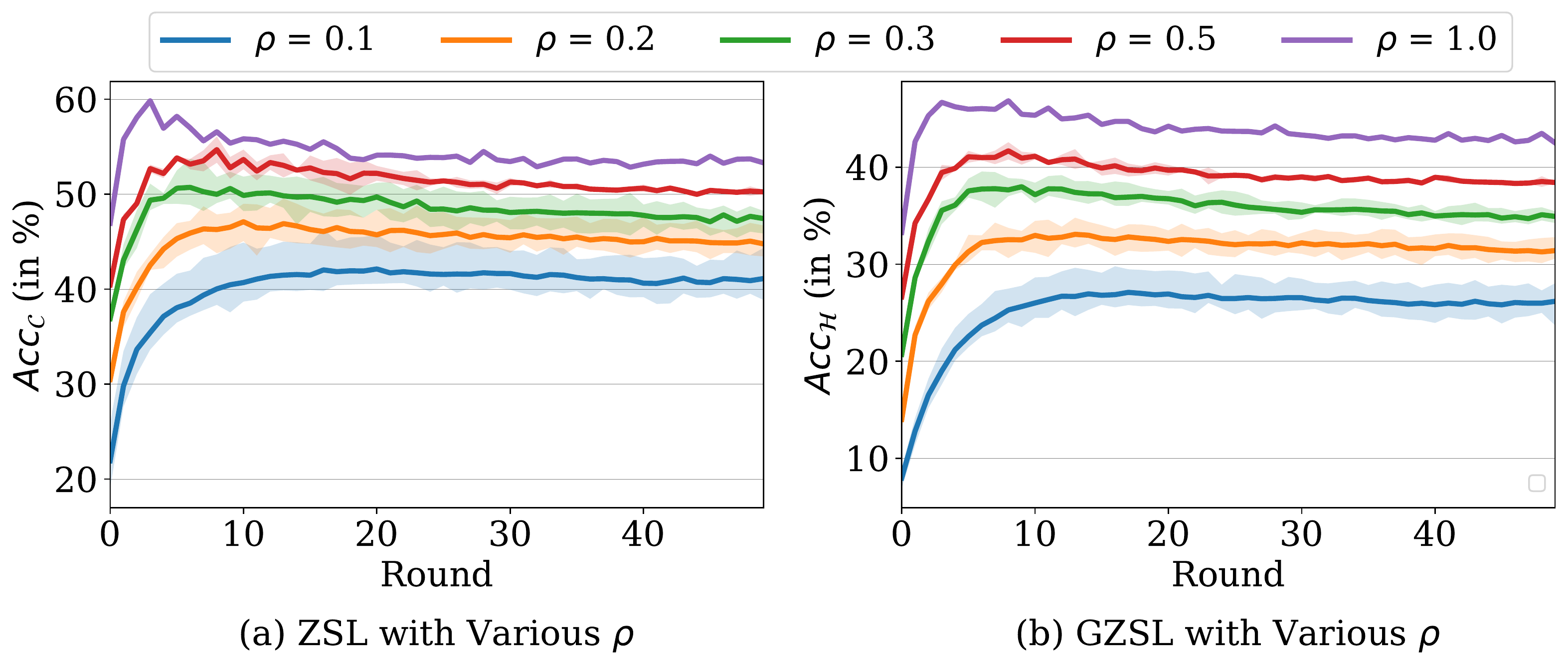}
    \caption{Learning curves of decentralized settings with partial data samples on the CUB dataset w.r.t the performance on ZSL and GZSL.}
    \label{fig:lesssamples}
    % \vspace{-25pt}
\end{figure}

\smallskip
\noindent \textbf{Impact of Partial Data Samples.} 
To study the impact of data quantity in the local classes, we conduct experiments on the CUB dataset by decentralizing the data onto ten clients. Each client only has a subset of training samples of all classes. For simplicity, we define the local data ratio $\rho = \{0.1, 0.2, 0.3, 0.5, 1.0\}$ to indicate the percentage of the total training samples in each client. In Fig. \ref{fig:lesssamples}, we depict the detailed learning curves with various $\rho$. It can be seen that even if being trained with only 10\% data (only $\sim$ 6 samples per class in CUB), the local model achieves more than 40\% on ZSL accuracy. In comparison %
% \begin{wrapfigure}{r}{7.cm}
%     \centering
%     \vspace{-12pt}
%     \includegraphics[width=1\linewidth]{figures/lesssamples.pdf}
%     \caption{Learning curves of decentralized settings with partial data samples on the CUB dataset w.r.t the performance on ZSL and GZSL.}
%     \label{fig:lesssamples}
%     \vspace{-25pt}
% \end{wrapfigure}
to the experiments on \texttt{Baselines} with various class percentage ratios, the improvements on training with more data are relatively lower. Also, we observe that there are only slight performance variances across devices, whereas the variances in \texttt{Baselines} are significant. This phenomenon demonstrates that the variation of data samples causes less impact than the variety of classes.

\section{Supplementary Backdoor Details}
In this section, we elaborate more on the backdoor attacks for testifying the robustness of the proposed \texttt{FedZSL}. We formalize the general procedure of the backdoor attacks applied as follows. We consider that there is a compromised (malicious) client $C_k$ holds a dataset $\mathcal{D}^{mal} = \{\vx^{mal}_i\}^r_{i=1}$ of size $r$ that is expected to be classified as the target labels $\{ y_i \}_{i=1}^r$. Let $\Gamma(\cdot)$ be the complete zero-shot classifier that predicts the label of the semantic attributes in $\mathcal{A}$, which has the highest compatibility score with $\vx_i$. Therefore, the local objective for client $C_k$ of the backdoor attacks is defined as:
\begin{equation}
\begin{aligned}
\mathcal{M}(\mathcal{D}^{s,k}\cup\mathcal{D}^{mal}, w_t^k) = \underset{w_t^k}{\max} \sum_{i=1}^{r} \mathbf{1}[\Gamma(\vx_i^{mal},\mathcal{A};w_t^k)=y_i].
\end{aligned}
\end{equation}
% Existing defensive methods are either inefficient or  in detecting the suspicious samples, \textit{e.g.,} gradient norm regularization \cite{sun2019can} needs calculating the gradients for each training sample.
% \noindent \textbf{Results with Disconnected Participants.} 
% \noindent \textbf{Supplementary Implementation Details.} 

As for the implementation of backdoor attacks, we start to involve the malicious samples in training when the global model is close to convergence. This injection strategy is proved to be more effective as indicated in \cite{bagdasaryan2020backdoor}. The backdoors injected in the very early rounds tend to be forgotten quickly. In particular, all the backdoor experiments start in communication round 20, where we observe that the global model tends to converge.

The backdoor samples $\vx^{mal}$ are injected by replacing the targeted training samples $\vx$ in each training batch. The probability of the replacement is set to 0.5. If the probability is higher, the main task performance can be affected. In contrast, if the probability is too low, the backdoor effects will be easily canceled out by the global aggregation. Moreover, to ensure the local model can generalize well to the malicious samples, each malicious sample to be injected is randomly sampled from $\mathcal{D}^{mal}$. Furthermore, to help the model generalize to the backdoor functionality, following \cite{chen2017targeted}, we add Gaussian noise with $\sigma=0.05$ to the malicious samples.

% The sampling strategy within the backdoor samples is random,  with a probability of 0.5. 

%, which is shown to be effective. 
% A higher replacing rate will cause performance drop for main tasks and a lower one will fail to inject.
% To tamper the training samples, following \cite{bagdasaryan2020backdoor} we start to randomly replace the training samples with the malicious samples when the 
% For GZSL performance, we adopt the Calibrated Stacking \cite{chao2016empirical} as a convention to balance the performance between the seen and unseen classes, which is implemented by reducing the prediction confidence of the output logits on seen classes.

\begin{figure}
    \centering
    % \vspace{-8pt}
    % \vspace{7pt}
    \includegraphics[width=0.95\linewidth]{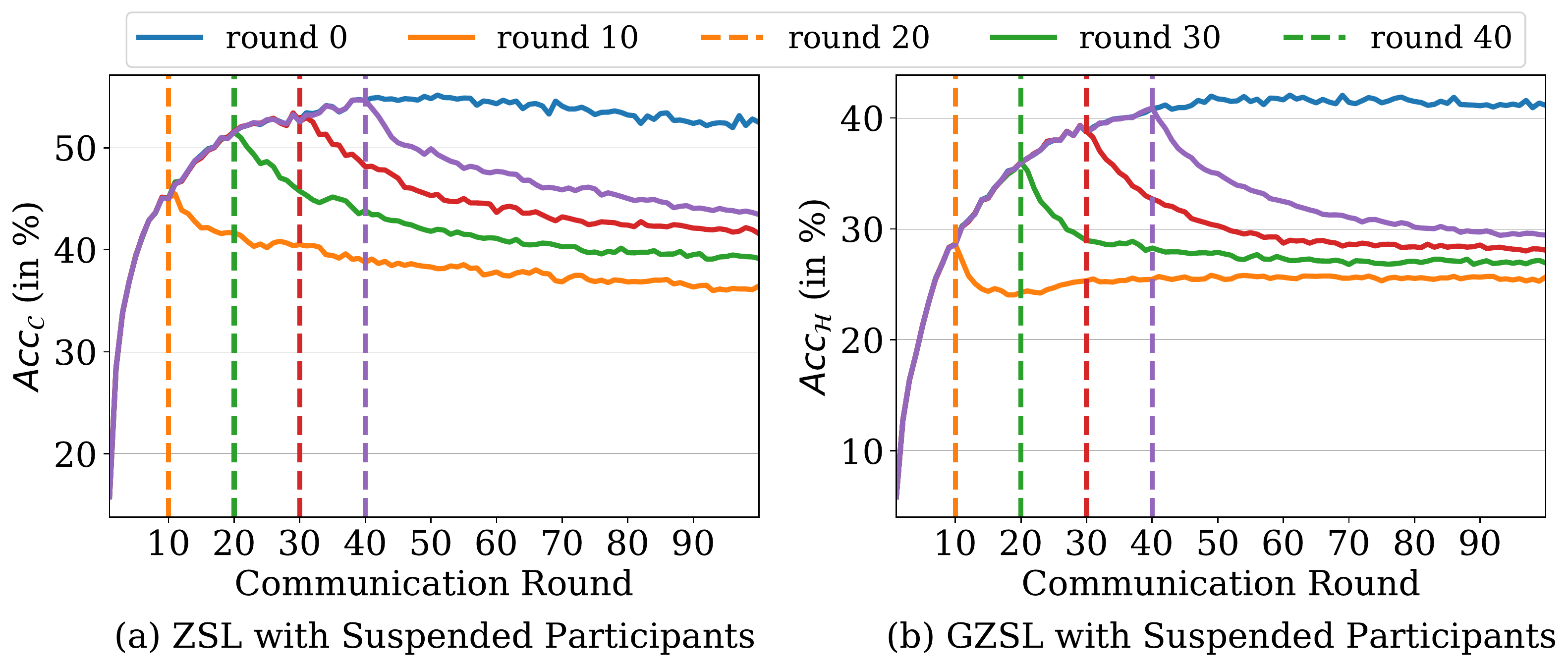}
    % \vspace{2.5pt}
    \caption{ZSL and GZSL learning curves with half participants being suspended in various communication rounds.}
    \label{fig:remove}
    % \vspace{-20pt}
\end{figure}

\section{Discussion of Limitations}
As is the case with a novel problem investigated and a general framework proposed, our work suggests many more questions than answers. Below, we discuss mainly two limitations that are not yet touched in this work, while being worth investigating to improve FedZSL in a near future. 

\smallskip
\noindent \textbf{Catastrophic Forgetting.}
First, we observe that when some participants are suspended from federated training, their contributions to the global model start to be gradually forgotten. As shown in Fig. \ref{fig:remove}, we conducted experiments with half participants (5/10) being suspended from federated training in the communication round 10, 20, 30, and 40. It can be seen that the performance significantly drops with the communication round goes up. We refer to this problem as catastrophic forgetting \cite{kirkpatrick2017overcoming} in \texttt{FedZSL}. This problem is also pervasive in continual learning \cite{hu2018overcoming}, in which a stream of incrementally arriving inputs is processed without access to the past data. Compared to continual learning, the catastrophic forgetting challenge in our learning scenario  is more severe considering the decentralized training scheme.  The problem statement can be described as follows. 
% \begin{wrapfigure}{r}{7.cm}
%     \centering
%     \vspace{-8pt}
%     % \vspace{7pt}
%     \includegraphics[width=0.95\linewidth]{figures/lessclasses.pdf}
%     % \vspace{2.5pt}
%     \caption{ZSL and GZSL learning curves with half participants being suspended in various communication rounds.}
%     \label{fig:remove}
%     \vspace{-20pt}
% \end{wrapfigure}
Given a sequence of federated learning stages or rounds in \texttt{FedZSL}, in which the training classes that appeared in the preceding learning stages may be absent in the subsequent learning stages. We aim to globally aggregate the local updates without accuracy deterioration that is caused by forgetting the contributions from the absent seen classes.

\smallskip
\noindent \textbf{Performance Degeneration.}
The second limitation we identify is the performance degeneration when training with an excess of rounds, which could be triggered by overfitting the seen classes. In Fig. \ref{fig:remove}(\textcolor{red}{a}), the blue solid line represents the learning curves of ZSL in 100 communication rounds. The default communication round number in this paper is set to 50, where the global model is converged. With an excess of communication rounds up to 100, we can clearly see the minor degeneration of the performance. Therefore, we argue that preventing the global model from overfitting to the local seen classes is a critical need for further improving the FedZSL framework.

\section{Conclusion }
In this work, we investigate a novel \textit{federated zero-shot learning} (\texttt{FedZSL}) problem by applying the CARD strategy to reach the consensus of visual-semantic predictions across devices and the FMD method for preventing backdoor attacks. Experiments validate that the proposed scheme is capable of handling new participants, various sampling fractions, and imbalanced class distribution. The proposed FMD is valid to stabilize model training and identify the malicious samples from various types.
% In this paper, we propose a novel problem setting of federated zero-shot learning (\texttt{FedZSL}) with the Cross-device Alignment via Relation Distillation (CARD) traning strategy and Feature Magnitude Defense (FMD) approach for defending backdoor attacks. Extensive experiments show that CARD effectively deals with \textit{p.c.c.d} data distribution and FMD can reduce the backdoor accuracy substantially with a minimal degradation in main tasks. 
% We identify two limitations in this paper. First, when a portion of participants quit from federated training, their contributions to the global model are canceled out after a few communication rounds. Second, we observe the phenomenon that when the server aggregates an excess of rounds, the performance start to degenerate. More discussions are provided in Appendix.

%%%%%%%%% REFERENCES
{\small
\bibliographystyle{ieee_fullname}
\bibliography{egbib}
}

\end{document}